\documentclass[journal]{IEEEtran}
\ifCLASSINFOpdf
\else
\fi

\usepackage{url}
\usepackage{booktabs}
\usepackage{amsmath}
\usepackage{amssymb}
\usepackage{dsfont}
\usepackage{graphicx}
\usepackage[caption=false,font=footnotesize]{subfig}
\usepackage{threeparttable}
\usepackage{makecell,multirow}
\usepackage{algorithm} 
\usepackage{algpseudocode} 
\usepackage{stfloats}
\usepackage{xcolor}
\usepackage{amsthm}

\newtheorem{theorem}{Theorem}
\newtheorem{lemma}{Lemma}
\theoremstyle{definition}

\newtheorem{assumption}{Assumption}
\newtheorem{corollary}{Corollary}
\theoremstyle{remark}
\usepackage{bbm}
\usepackage{amssymb}    % for \checkmark
\usepackage{pifont}     % for \xmark
\newtheorem{remark}{Remark}
\newcommand{\xmark}{\ding{55}}  % 对应叉号 ✗6
\newcommand{\review}[1]{{\color{black}{#1}}}
\begin{document}
%
% paper title
% Titles are generally capitalized except for words such as a, an, and, as,
% at, but, by, for, in, nor, of, on, or, the, to and up, which are usually
% not capitalized unless they are the first or last word of the title.
% Linebreaks \\ can be used within to get better formatting as desired.
% Do not put math or special symbols in the title.
\title{\huge Deep Fictitious Play-Based Potential Differential Games for Learning Human-Like Interaction at Unsignalized Intersections}
%
%
% author names and IEEE memberships
% note positions of commas and nonbreaking spaces ( ~ ) LaTeX will not break
% a structure at a ~ so this keeps an author's name from being broken across
% two lines.
% use \thanks{} to gain access to the first footnote area
% a separate \thanks must be used for each paragraph as LaTeX2e's \thanks
% was not built to handle multiple paragraphs
%

\author{Kehua~Chen,~\IEEEmembership{Member, IEEE}, Ryan Feng Lin,~\IEEEmembership{Graduate Student Member, IEEE}, Shucheng~Zhang, Yinhai~Wang,~\IEEEmembership{Fellow, IEEE}
\thanks{Kehua Chen, Shucheng Zhang, and Yinhai Wang are with the Department of Civil and Environmental Engineering, University of Washington, Seattle, United States.}% <-this % stops a space
\thanks{Ryan Feng Lin is with the Department of Industrial \& Systems Engineering, University of Washington, Seattle, United States.}
\thanks{*Yinhai Wang is the corresponding author, E-mail: yinhai@uw.edu}
}

% The paper headers
\markboth{Journal of \LaTeX\ Class Files}%
{Shell \MakeLowercase{\textit{et al.}}: Bare Demo of IEEEtran.cls for IEEE Journals}

\maketitle

% As a general rule, do not put math, special symbols or citations
% in the abstract or keywords.
\begin{abstract}
Modeling vehicle interactions at unsignalized intersections is a challenging task due to the complexity of the underlying game-theoretic processes. Although prior studies have attempted to capture interactive driving behaviors, most approaches relied solely on game-theoretic formulations and did not leverage naturalistic driving datasets. In this study, we learn human-like interactive driving policies at unsignalized intersections using Deep Fictitious Play. Specifically, we first model vehicle interactions as a Differential Game, which is then reformulated as a Potential Differential Game. The weights in the cost function are learned from the dataset and capture diverse driving styles. We also demonstrate that our framework provides a theoretical guarantee of convergence to a Nash equilibrium. To the best of our knowledge, this is the first study to train interactive driving policies using Deep Fictitious Play. We validate the effectiveness of our Deep Fictitious Play-Based Potential Differential Game (DFP-PDG) framework using the INTERACTION dataset. The results demonstrate that the proposed framework achieves satisfactory performance in learning human-like driving policies. The learned individual weights effectively capture variations in driver aggressiveness and preferences. Furthermore, the ablation study highlights the importance of each component within our model. Code is available at \url{https://github.com/zeonchen/DFP-PDG}.
\end{abstract}

% Note that keywords are not normally used for peerreview papers.

% For peer review papers, you can put extra information on the cover
% page as needed:
% \ifCLASSOPTIONpeerreview
% \begin{center} \bfseries EDICS Category: 3-BBND \end{center}
% \fi
%
% For peerreview papers, this IEEEtran command inserts a page break and
% creates the second title. It will be ignored for other modes.
\IEEEpeerreviewmaketitle

\section{Introduction}

Unsignalized intersections are common scenarios characterized by strong interactive driving behaviors. Previous studies have made significant efforts to model the complex interactions among vehicles in such environments. For instance, \cite{hang2022decision} and \cite{hang2022driving} proposed decision-making frameworks for connected automated vehicles (CAVs) based on fuzzy coalitional games and differential games, respectively. Similarly, \cite{liu2024cooperative} formulated the decision-making process as a decentralized multi-agent reinforcement learning problem, incorporating game-theoretic priors to model interaction behaviors. To address environmental noise during driving, \cite{huang2024non} introduced a robust differential game framework capable of generating both cooperative and non-cooperative strategies. Nonetheless, these studies primarily derived driving policies from predefined environments or simulations, which limits their ability to capture authentic human interactions. To derive human-like driving strategies, some studies have leveraged naturalistic datasets or incorporated diverse driving styles into their frameworks \cite{jing2024decentralized,wang2023learning,hu2024modeling}.

However, there are still some limitations to these studies.
First, although many studies employed game-theoretic methods, they primarily produced high-level and discrete decisions (e.g., yield/go or enter/hold at the conflict zone) or coarse motion primitives.
Such outputs must be translated into low-level continuous controls subject to vehicle kinematic/dynamic constraints, which is often nontrivial and may lead to physically infeasible or overly aggressive trajectories without additional hand-crafted control modules.
Second, most previous studies manually designed cost functions and assigned weights, and the resulting behavior can be highly sensitive to these weights.
In practice, this often requires scenario-specific calibration or extensive tuning (e.g., grid search), limiting both the expressiveness of the model and its generalization across different intersection geometries, traffic densities, and driving styles.
Third, many existing approaches lack an explicit dynamic mechanism for evolving interaction strategies among agents. They typically compute one-shot plans or treat opponents as fixed predictions.
As a result, they may fail to capture the mutual adaptation and negotiation process (e.g., creeping, yielding, and sequential adjustments) that is crucial in unsignalized intersections.

To address the aforementioned limitations, we propose a Deep Fictitious Play-Based Potential Differential Game (DFP-PDG) in this study. To the best of our knowledge, this is the first study to employ Deep Fictitious Play (DFP) for deriving interactive driving strategies. Specifically, we first model the interactions among vehicles at unsignalized intersections as a differential game, and then reformulate it as a potential differential game. In addition, we interpret the weights in the weighted potential game as latent driving-style parameters, which enables the framework to represent heterogeneous behaviors across drivers within a unified game-theoretic structure. Subsequently, the driving policy is learned through a deep policy network based on the naturalistic dataset. Afterwards, we adopt a differentiable optimization framework, where the initial solution proposed by the deep policy is refined via a structured optimization layer. This enables gradient-based learning through both the network and the optimization process. The framework is trained using DFP, wherein the policy of one agent is optimized while keeping the policies of the other agents fixed. At the end of the methodology section, we provide a theoretical analysis showing that the underlying cyclic best-response dynamics in our weighted potential formulation admits a descent interpretation on a single global potential and converges to a (local) Nash equilibrium under standard regularity assumptions. In the experimental section, we evaluate the effectiveness of our model using the INTERACTION dataset. A detailed analysis of the learned weights and the ablation study further support the validity and effectiveness of the proposed method.

To sum up, the main contributions of this study include:

\begin{itemize}
    \item \textbf{Framework design.}
    We propose a Deep Fictitious Play-Based Potential Differential Game (DFP-PDG) framework that couples a deep policy network that provides a warm start and learnable heterogeneity weights with a differentiable nonlinear optimization layer that computes an approximate best response, enabling end-to-end learning through the best-response refinement in continuous multi-agent interactions.

    \item \textbf{Theory.}
    We reformulate unsignalized-intersection interactions from a general differential game into a \emph{weighted potential differential game}, which makes fictitious-play updates interpretable as approximate block-coordinate descent on a single global potential. Under standard assumptions, we provide convergence results to a Nash equilibrium for exact best responses, and to an $\varepsilon$-Nash equilibrium under bounded approximation errors.

    \item \textbf{Data-driven representation and heterogeneity modeling.}
    We design a multi-modal deep policy network that integrates rasterized semantic maps, goal states, and multi-agent historical trajectories to learn human-like behaviors from naturalistic data, and jointly predicts motion controls together with per-driver style weights that capture heterogeneous preferences within a unified game-theoretic structure.

    \item \textbf{Empirical validation and interpretability.}
    We validate the proposed framework on two real-world unsignalized-intersection scenarios from the INTERACTION dataset, demonstrating competitive ADE/FDE and improved safety behavior compared to strong baselines. We further analyze the learned heterogeneity weights and conduct ablations to verify the role of each component in accuracy, safety, and stability.
\end{itemize}

The remainder of the paper is organized as follows. We briefly introduce related works in Section 2. Afterwards, the proposed method is presented in Section 3. Section 4 thoroughly introduces experiments for evaluation, results, and discussions. At last, we summarize the overall paper in Section 5.

\section{Related Work}
\subsection{Human Driver Behavior Modeling}
Recent research has focused on learning driving policies directly from naturalistic datasets to consider human factors in behavior modeling. In this section, we primarily review studies that aim to achieve human-like driving.
In highway scenarios, \cite{xu2022driving} proposed an integrated driving behavior model that makes decisions in a step-wise manner. \cite{xu2020learning} recovered human driving cost functions using a softmax transformation, which were subsequently employed for trajectory selection. \cite{wang2022high} applied behavior cloning to learn human driving behaviors from naturalistic datasets, with a focus on high-level decision-making. Additionally, \cite{xia2021human} introduced the Human-like Lane Changing Intention Understanding Model (HLCIUM), which leverages a Hidden Markov Model (HMM) to infer the lane-changing intentions of surrounding vehicles. \cite{lu2023human} first constructed a cognitive map using successor representations derived from human driving datasets and then performed decision-making based on a motion primitive library to achieve human-like driving behavior. 

Nevertheless, these studies did not explicitly account for the interactive dynamics between vehicles. To model vehicle interactions, \cite{zhao2024human} first employed Inverse Reinforcement Learning (IRL) to recover structured cost functions that account for social value orientation. High-level decisions were subsequently made using a Stackelberg game framework, followed by Model Predictive Control (MPC) for motion planning.
\cite{chen2024combining} proposed a Deep Markov Cognitive Hierarchy Model to capture the game-theoretic decision-making process among vehicles during lane-changing maneuvers. 
\cite{chen2024human} proposed a framework, Diff-LC, for human-like lane-changing planning based on diffusion models. To capture interactions between vehicles, the authors incorporated Multi-Agent Adversarial Inverse Reinforcement Learning (MA-AIRL) to evaluate the generated trajectories. 

For intersection scenarios, \cite{jing2024decentralized} modeled vehicle interactions as a Stackelberg game and employed a driving style recognition algorithm to solve the game using distinct cost functions. However, the Stackelberg game framework is limited to generating only discrete decisions. Similarly, \cite{wang2023learning} proposed a behavior cloning model to learn safe and human-like high-level driving decisions from naturalistic datasets. The model also demonstrated strong interpretability. \cite{hu2024modeling} proposed a human-like driving model for signalized intersections based on a driver risk field framework. The approach incorporated predefined rules and cost functions to emulate human-like behavior; however, the model required plenty of rules and evaluation was conducted solely in simulation. \cite{shu2023human} applied a linear quadratic differential game to model left-turning behavior at intersections, using dynamic time-to-collision (TTC) to represent the driving styles of interacting vehicles. Nonetheless, the ego vehicle's policy was still derived from hand-crafted utility functions.

Specifically, our study is closely related to two prior works. First, \cite{diehl2023energy} proposed an Energy-Based Potential Games (EPO) framework for vehicle trajectory prediction. EPO is trained end-to-end by jointly learning a neural game-parameter inference module and a differentiable game-theoretic optimization layer, but it does not explicitly model iterative policy updates. In contrast, we adopt a DFP-style iterative best-response procedure to learn interactive driving policies, and we further provide convergence guarantees under a weighted potential differential-game formulation, highlighting the role of learnable heterogeneity weights for capturing diverse driving styles.

Additionally, DIPP \cite{huang2023differentiable} proposed a fully differentiable integrated prediction–planning pipeline, in which a differentiable nonlinear optimizer serves as the ego-vehicle planner and the associated cost weights are learned from data. Importantly, DIPP primarily targets single-agent planning and does not formulate training as multi-agent fictitious play or equilibrium learning, nor does it establish DFP-style convergence results under a potential-game structure. In contrast, our method explicitly models interaction and learns interactive policies, leading to a fundamentally different theoretical object and algorithmic pathway.

\subsection{Fictitious Play}
Fictitious Play (FP) models opponents as playing stationary mixed strategies \cite{rescorla2007convention}. Each player forms beliefs from empirical frequencies of observed past actions and plays a best response to those beliefs. For traditional FP, \cite{robinson1951iterative} established core convergence properties of FP toward mixed Nash equilibrium behavior. \cite{monderer1996potential} formalized potential games and showed broad classes enjoy the fictitious play property.

Classical FP is iterative in finite-action games, but in differential games (DGs) with continuous states/actions, each best response is a high-dimensional stochastic control problem governed by a partial differential equation. \cite{hu2019deep} introduced DFP to compute open-loop Nash equilibria in multi-player stochastic differential games. The method follows the classical fictitious-play template. The authors further parameterized open-loop controls with neural networks, and provided convergence results to an open-loop equilibrium in linear–quadratic settings. Moreover, \cite{han2020convergence} developed DFP for Markovian Nash equilibria in N-player stochastic differential games, decoupled the game into parallel single-agent control problems, and solved them via deep BSDE methods. However, these studies mainly address general differential games under restrictive technical conditions. They do not explicitly use potential-structure descent, nor do they offer equally direct guarantees for cyclic, approximate-update training and equilibrium selection, which are important in interactive driving.

On the other hand, potential games formalize settings where each player’s unilateral improvement is tracked by a single global potential, which makes best-response and fictitious-play–style learning particularly well matched to equilibrium computation \cite{monderer1996potential}. In the continuous setting, a fictitious-play procedure is shown to converge for potential mean-field games, clarifying when iterative “learn-and-best-respond” dynamics are stable at scale \cite{cardaliaguet2017learning}. However, that study focuses on mean-field games with a large population of agents, rather than interactive driving scenarios with a small number of strategically coupled participants.

\section{Methodology}
\subsection{Problem Formulation}
\begin{figure}[htbp]
  \centering
  \includegraphics[width=0.85\linewidth]{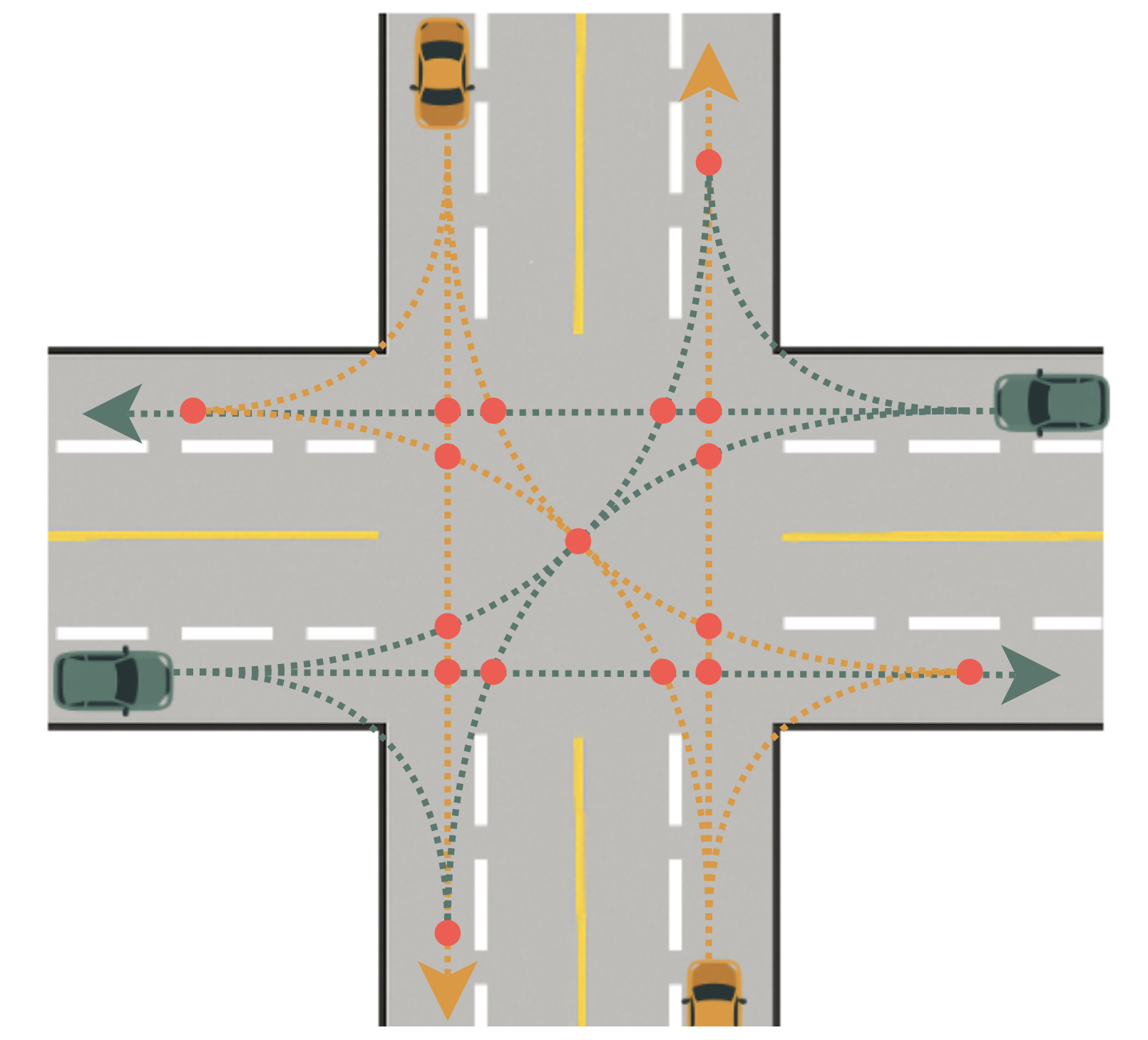}
  \caption{The conflict points at an unsignalized intersection.}
  \label{fig:interacting}
\end{figure}

At an unsignalized intersection, the conflict points can be identified as shown in Fig. \ref{fig:interacting}. 
Unlike previous studies that primarily modeled interactions between two vehicles, our study extends scenarios involving interactions among multiple vehicles.

In this study, we focus on vehicle motion planning at unsignalized intersections.
\review{Formally, assume we have $N$ interacting vehicles denoted by $\mathcal{N}=\{1,2,\cdots, N\}$. For each ego vehicle $i\in \mathcal{N}$, we would like to plan human-like future motions, given its goal state (i.e., the target position) $\mathbf{g}_i \in \mathbb{R}^5$ and the historical trajectories of all the vehicles of length $T_h$. 

At each time $t$, the state of the ego vehicle $i$ is denoted by $\mathbf{s}_i(t)\in \mathcal{S}$, where $\mathcal{S}\subseteq\mathbb{R}^5$ is the full state space spanning across five dimensions: velocities $v_x, v_y$, locations $x, y$, and heading $\psi$. To avoid ambiguity and ensure physical executability, we denote the two-dimensional control/action vector at time $t$ by $\mathbf a_i(t)\in\mathcal{A}$ where $\mathcal{A}\subseteq\mathbb R^2$ is the full action space. In our implementation, rather than a free two-dimensional acceleration vector, this action is instantiated as standard kinematically feasible controls, i.e., longitudinal acceleration and steering angle:
\[
\mathbf a_i(t) = [a_i^{\mathrm{lon}}(t), \delta_i(t)]^\top .
\]
Given the $i$-th vehicle's last state $\mathbf s_i(t-1)$ and its next action $\mathbf a_i(t)$ by a policy function $\pi_i\in \Pi:\mathcal{S}\to \mathcal{A}$, we have $\mathbf a_i(t) =\pi_i(\boldsymbol{s}_i(t-1))$, where $\Pi$ is the space of all feasible policy functions. Through a unicycle model, this action can result in an updated state. Thus, starting from the initial state $\mathbf s_i(0)$, the future state-action trajectory up to time $T_f$ is generated by a deterministic unicycle rollout:
\[
\boldsymbol{\tau}_i(T_f) = [\mathbf s_i(0), \mathbf a_i(1),\mathbf s_i(1),\cdots,\mathbf a_i(T_f-1),\mathbf s_i(T_f)],
\]
where the state evolution in $\boldsymbol{\tau}_i$ implicitly encodes necessary information of positions, velocities, and heading over the planning horizon. These quantities are not fully independent decision variables, but can be induced by $\mathbf a_i$.}

To model the complex interaction behaviors among vehicles, we formulate the problem as a differential game \cite{friedman2013differential}. The joint state and actions of $N$ vehicles at time $t$ are represented by:
\begin{align}
    \mathbf{s}(t) = [\mathbf{s}_1(t),...,\mathbf{s}_{N}(t)]\\
    \mathbf{a}(t) = [\mathbf{a}_1(t),...,\mathbf{a}_{N}(t)]
\end{align}
\review{which results in a joint set of their trajectories:
\begin{align}
   \boldsymbol{\tau}(t)=[\boldsymbol{\tau}_1(t),\cdots,\boldsymbol{\tau}_N(t)] 
\end{align}
As a result, each vehicle determines its own policy function $\pi_i$ to alter its rollout trajectory to minimize its own cost ${J}_i$. With a little abuse of the notation, the individual cost within the time interval $[0, T_f]$ can be written as:
\begin{align}
\begin{aligned}
    {J}_i(\pi_1,\cdots, \pi_N) &= J^{(i)}(\boldsymbol{\tau}(T_f))\\
    &=\sum_{t=1}^{T_f} c_i(\mathbf{s}(t),\mathbf{a}(t)) \;+\; \phi_i(\mathbf{s}(T_f)).
\end{aligned}
\label{Jdef}
\end{align}}
where $c_i(\cdot, \cdot)$ is the stage cost function measuring the instantaneous cost during driving, and $\phi(\cdot)$ is the terminal cost measuring the cost at the end of the time interval. \review{In our paper, we assume all these cost components are continuous functions.}

We design $c_i$ and $\phi_i$ to reflect goal-reaching, smoothness, efficiency, and safety.
The terminal cost penalizes terminal goal error with a regularizer $\lambda_{\text{goal}}>0$:
\begin{align}
\phi_i(\mathbf{s}_i(T_f))
\triangleq \lambda_{\text{goal}}\big\|\mathbf{s}_i(T_f)-\mathbf{g}_i\big\|_2^2.
\label{eq:phi_def}
\end{align}
The stage cost can be decomposed into an individual term and an interaction term:
\review{
\begin{align}
c_i(\mathbf{s}(t),\mathbf{a}(t))
&\triangleq c_{i,\text{ind}}(t) \;+\; c_{i,\text{safety}}(t),
\label{eq:ci_split}\\
c_{i,\text{ind}}(t)
&\triangleq \lambda_{\text{smooth}}
\big\|\bar{\mathbf{a}}_i(t)-\bar{\mathbf{a}}_i(t-1)\big\|_2^2 \nonumber\\
&\quad + \lambda_{\text{efficiency}}
\big(v_i^{\mathrm{prog}}(t)-v_i^{\mathrm{ref}}\big)^2,
\label{eq:ci_ind}
\\
c_{i,\text{safety}}(t)
&\triangleq \lambda_{\text{safety}}\sum_{j\neq i} f_{ij}(t),
\label{eq:ci_safety}
\end{align}
where $\lambda_{\text{smooth}}$, $\lambda_{\text{efficiency}}$, and $\lambda_{\text{safety}}$ are positive constants that balance the relative importance of different components. 

For Eq. \ref{eq:ci_ind}, $\bar{\mathbf{a}}_i(t)$ denotes the normalized control vector, since longitudinal acceleration and steering angle have different physical units.
The first term penalizes rapid control variations, including sudden changes in longitudinal acceleration and steering.
The second term measures driving efficiency through the rollout-induced progress speed rather than directly rewarding large control magnitudes.
Specifically, let $\mathbf{p}_i(t)\in\mathbb{R}^2$ and $\dot{\mathbf{p}}_i(t)\in\mathbb{R}^2$ denote the position and velocity of agent $i$ at time $t$, we define
\[
v_i^{\mathrm{prog}}(t)
=
\mathbf r_i(t)^\top \dot{\mathbf p}_i(t),
\quad
\mathbf r_i(t)
=
\frac{\mathbf g_i^{p}-\mathbf p_i(t)}
{\|\mathbf g_i^{p}-\mathbf p_i(t)\|_2+\varepsilon},
\]
where $\mathbf g_i^{p}$ is the position component of the goal state and $\mathbf r_i(t)$ is the unit direction from the current position to the goal.
The reference speed $v_i^{\mathrm{ref}}$ is computed from the observed historical speed or the dataset-level nominal speed.
Thus, the efficiency term encourages reasonable forward progress toward the goal without encouraging unrealistically large acceleration or steering inputs.}

For Eq. \ref{eq:ci_safety}, $f_{ij}(t)=f_{ji}(t)$ is a symmetric pairwise safety surrogate constructed from relative position and relative velocity. Define
\begin{align}
d_{ij}(t) &\triangleq \left\|\mathbf{p}_i(t)-\mathbf{p}_j(t)\right\|_2, \nonumber\\
u_{ij}(t) &\triangleq -\frac{\big(\mathbf{p}_i(t)-\mathbf{p}_j(t)\big)^{\!\top}
\big(\dot{\mathbf{p}}_i(t)-\dot{\mathbf{p}}_j(t)\big)}{d_{ij}(t)+\varepsilon}, \nonumber\\
\beta_{ij}(t) &\triangleq 1+\alpha\,\sigma\!\left(\frac{u_{ij}(t)-u_0}{\sigma_u}\right), \nonumber\\
f_{ij}(t)
&\triangleq \beta_{ij}(t)\,
\operatorname{sp}\!\left(\frac{d_{\text{safe}}-d_{ij}(t)}{\sigma_d}\right)^{2},
\label{eq:fij_def}
\end{align}
where $d_{ij}(t)$ is the inter-agent distance; $u_{ij}(t)$ is the \emph{closing rate} (positive when agents are approaching and near zero/negative when separating); $\varepsilon>0$ is a small constant to avoid division by zero.
The function $\sigma(\cdot)$ denotes the sigmoid, and $\mathrm{sp}(x)=\log(1+e^x)$ is the softplus.
In the gating weight $\beta_{ij}(t)$, $\alpha\ge 0$ controls the strength of closing-rate upweighting, $u_0$ is the closing-rate threshold at which the gate starts to activate, and $\sigma_u>0$ controls the smoothness (steepness) of the sigmoid transition.
In the distance barrier, $d_{\text{safe}}>0$ is the desired safety margin and $\sigma_d>0$ controls the softness of the barrier.
Overall, $\beta_{ij}(t)$ upweights near-collision cases with large closing rate and downweights benign close-pass events where the distance is increasing.

To solve the differential game, we may find a joint strategy $\{\pi_i^*\}_{i=1}^{N}$ that is a Nash equilibrium for each $i\in \mathcal{N}$:
\begin{align}
{J}_i(\pi_i^*,\pi_{-i}^*)\leq {J}_i(\pi_i,\pi_{-i}^*), \forall \mathbf{a}_i
\end{align}
where $\pi_{-i}^*$ denotes the policies of all other vehicles except $i$.

\subsection{From Differential Game to Potential Differential Game}
Although general differential games can be effectively modeled in multi-vehicle interactions, they often suffer from high computational complexity and are challenging to integrate with modern learning-based models \cite{fonseca2018potential, varga2024identification}. By transforming a differential game into a potential differential game, we can rely on the \emph{structural property} of our interaction objective to simplify the optimization with a shared global potential. Without altering the system dynamics or feasibility constraints, we show that the original differential game admits a potential representation.
Concretely, we construct a global potential function $\Phi$ and verify that any unilateral change of an individual vehicle's strategy induces a proportional change in its own objective, which is exactly the defining condition of a potential differential game.
This identification allows us to interpret alternating best-response updates as coordinate descent steps on $\Phi$, forming the basis of our convergence analysis.

\begin{figure}[!t]
  \centering
  \includegraphics[width=\linewidth]{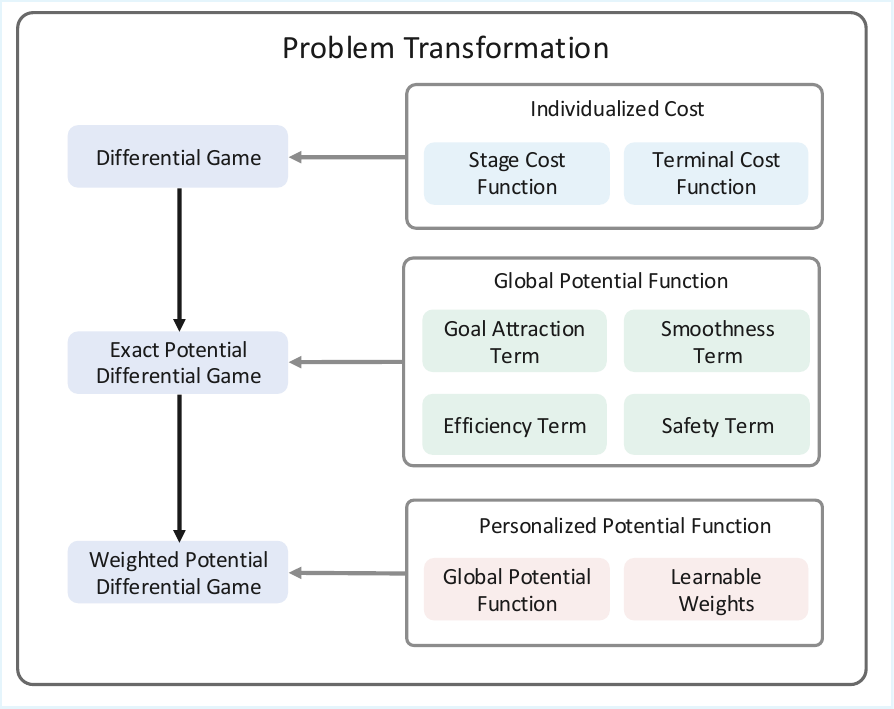}
  \caption{Problem transformation process in this study. The problem is first modeled as a differential game, then transformed into a potential game.}
  \label{fig:problem}
\end{figure}

In detail, we can construct a globally shared function by aggregating (i) all individual terms and (ii) the pairwise safety terms over unordered pairs:
\begin{align}
\begin{aligned}
\Phi(\boldsymbol{\tau}(T_f))\triangleq &\sum_{i=1}^{N}\phi_i(\mathbf{s}_i(T_f))
+ \sum_{t=1}^{T_f}\Bigg[\sum_{i=1}^{N} c_{i,\text{ind}}(t)\\
&+ \lambda_{\text{safety}}\sum_{i<j} f_{ij}(t)\Bigg].
\end{aligned}
\label{eq:Phi_def}
\end{align}
\review{Here, $\Phi$ aggregates the terminal, running and interaction costs over all vehicles. This permutation-invariant form ensures any unilateral change by a vehicle only affects the
corresponding terms. Now we show that the differential game with $\Phi$ is exactly a dynamic potential game \cite{guo2025towards} as follows:}

\review{\begin{theorem}[Exact Potential Game]
\label{prop:exact_potential_rollout}
Let $\Pi$ be a common admissible policy space for the $N$ vehicles. Let $\pi=(\pi_1,\cdots, \pi_N)\in\Pi^N$ denote the joint policy profile, and $\pi_{-i}=(\pi_1,\cdots,\pi_{i-1},\pi_{i+1},\cdots, \pi_N)\in\Pi^{N-1}$ denote the policies of all vehicles except $i$. Fix the initial states $\boldsymbol{s}_i(0)$ for $\forall i=1,\cdots, N$. Under the unicycle rollout, $\Phi$ is an exact dynamic potential function for the game. Specifically, for any vehicle $i\in\mathcal{N}$, any unilateral policy function $\pi_i\in\Pi$ and its deviation $\tilde{\pi}_i\in\Pi$ and fixed $\pi_{-i}\in\Pi^{N-1}$, we have:
\begin{align*}
% \begin{aligned}
{J}_i(\pi_i,\pi_{-i})-{J}_i(\tilde{\pi}_i,\pi_{-i}) 
=
\Phi(\boldsymbol{\tau}_i,\boldsymbol{\tau}_{-i})-\Phi(\tilde{\boldsymbol{\tau}}_i,\boldsymbol{\tau}_{-i}).
% \end{aligned}
\end{align*}
\end{theorem}
\begin{proof}
Consider the deterministic rollout trajectory $\boldsymbol{\tau}_i(T_f)$ for vehicle $i$. Then, with Eq. \eqref{eq:Phi_def}, we have
\begin{align*}
&\Phi(\boldsymbol{\tau}_i(T_f), \boldsymbol{\tau}_{-i}(T_f)) - \Phi(\tilde{\boldsymbol{\tau}}_i(T_f), \boldsymbol{\tau}_{-i}(T_f))\\
=&\Bigl[\phi_i(\mathbf{s}_i(T_f))-\phi_i(\tilde{\mathbf{s}}_i(T_f))\Bigr]+\sum_{t=1}^{T_f}\Bigl[\Bigl(c_{i,\text{ind}}(t) - \tilde{c}_{i,\text{ind}}(t)\Bigr)\\
&+\lambda_{\text{safety}}\Bigl(\sum_{j>i }\bigl(f_{ij}(t)-\tilde{f}_{ij}(t)\bigr)+\sum_{j<i }\bigl(f_{ji}(t)-\tilde{f}_{ji}(t)\bigr)\Bigr)\Bigr]
\end{align*}
Due to the symmetry property $f_{ij}(t)=f_{ji}(t)$, we have $\tilde{f}_{ij}(t)=\tilde{f}_{ji}(t)$. Together with Eqs. \eqref{eq:ci_split} and \eqref{eq:ci_safety}, we have 
\begin{align*}
&\Phi(\boldsymbol{\tau}_i(T_f), \boldsymbol{\tau}_{-i}(T_f)) - \Phi(\tilde{\boldsymbol{\tau}}_i(T_f), \boldsymbol{\tau}_{-i}(T_f))\\
=&\Bigl[\phi_i(\mathbf{s}_i(T_f))-\phi_i(\tilde{\mathbf{s}}_i(T_f))\Bigr]+\sum_{t=1}^{T_f}\Bigl[\Bigl(c_{i,\text{ind}}(t) - \tilde{c}_{i,\text{ind}}(t)\Bigr)\\
&+\lambda_{\text{safety}}\sum_{j\neq i}\Bigl(f_{ij}(t)-\tilde{f}_{ij}(t)\Bigr)\Bigr]\\
=&\Bigl[\phi_i(\mathbf{s}_i(T_f))-\phi_i(\tilde{\mathbf{s}}_i(T_f))\Bigr]+\sum_{t=1}^{T_f}\Bigl[\Bigl(c_{i,\text{ind}}(t) - \tilde{c}_{i,\text{ind}}(t)\Bigr)\\
&+\Bigl(c_{i,\text{safety}}(t) - \tilde{c}_{i,\text{safety}}(t)\Bigr)\Bigr]\\
=&\Bigl[\phi_i(\mathbf{s}(T_f))-\phi_i(\tilde{\mathbf{s}}(T_f))\Bigr]\\
&+\sum_{t=1}^{T_f}\Bigl(c_i(\boldsymbol{s}(t),\mathbf{a}(t))-c_i(\tilde{\boldsymbol{s}}(t),\tilde{\mathbf{a}}(t))\Bigr)\\
=&J_i(\pi_i,\pi_{-i})-J_i(\tilde{\pi}_i,\pi_{-i})
\end{align*}
which completes the proof.
\end{proof}}
\review{
In practice, the policies are typically implemented through differentiable parameterizations, such as neural networks. Suppose the policy function of the vehicle $i$ parametrized by $\boldsymbol{\theta}_i\in\Theta$ as $\pi_{\boldsymbol{\theta}_i}$, where the parameter space $\Theta$ is open. We have the following property:

\begin{corollary}
Suppose the conditions of Theorem 1 hold. Assume that the unicycle rollout and all cost components are differentiable along the generated trajectories. For vehicle $i$, with any admissible direction $\boldsymbol{v}_i$, we have
\begin{align}
\frac{d}{d\varepsilon}{J}_i(\pi_{\boldsymbol{\theta}_i+\varepsilon\boldsymbol{v}_i},\pi_{-i})\Bigl|_{\varepsilon=0}=\frac{d}{d\varepsilon}\Phi(\boldsymbol{\tau}_i^{\varepsilon}(T_f),\boldsymbol{\tau}_{-i}(T_f))\Bigl|_{\varepsilon=0}.
\label{eq:grad_potential}
\end{align}
where $\boldsymbol{\tau}_i^{\varepsilon}$ is a perturbed rollout trajectory.
\end{corollary}
\begin{proof}
Fix the direction $\boldsymbol{v}_i$. Since $\Theta$ is open, it is easy to see $\boldsymbol{\theta}_i^{\varepsilon}=\boldsymbol{\theta}_i+\varepsilon\boldsymbol{v}_i\in\Theta$ for $\varepsilon\to 0$. Let $\tilde{\pi}_i=\pi_{\boldsymbol{\theta}_i+\varepsilon\boldsymbol{v}_i}$. Then we can easily see $\pi_i=\tilde{\pi}_i|_{\varepsilon=0}$. Let $U_i$ denote the unicycle model. Note that the unicycle rollout generates the actions and states recursively. When the policy $\pi_i$ is unilaterally changed to $\tilde{\pi}_i$ with $\boldsymbol{\theta}_i$ perturbed by $\varepsilon$ in $\boldsymbol{v}_i$, the perturbed state-action trajectory generated by the resulting rollout can be written as:
\begin{align*}
    \boldsymbol{\tau}_i^{\varepsilon}(T_f)=[\mathbf{s}_i^{\varepsilon}(0), \mathbf{a}^{\varepsilon}_i(1),\mathbf{s}_i^{\varepsilon}(1),\cdots, \mathbf{a}^{\varepsilon}_i(T_f-1),\mathbf{s}_i^{\varepsilon}(T_f)]
\end{align*}
where $\mathbf{s}_i^{\varepsilon}(0)=\mathbf{s}_i(0)$ and 
\begin{align*}
&\mathbf{a}_i^{\varepsilon}(t)=\pi_{\boldsymbol{\theta}^{\varepsilon}_i}(\mathbf{s}_i^{\varepsilon}(t-1)),\\
&\mathbf{s}_i^{\varepsilon}(t)=U_i(\mathbf{s}_i^{\varepsilon}(t-1), \mathbf{a}^{\varepsilon}_i(t))
\end{align*}
for $t=1,\cdots,T_f$. Then, by Theorem 1, we have
\begin{align*}
&{J}_i(\pi_{\boldsymbol{\theta}_i+\varepsilon\boldsymbol{v}_i},\pi_{-i})-{J}_i(\pi_{\boldsymbol{\theta}_i},\pi_{-i})\\
=&J_i(\tilde{\pi}_i,\pi_{-i}) - J_i(\pi_i,\pi_{-i})=\Phi(\tilde{\boldsymbol{\tau}}_i,\boldsymbol{\tau}_{-i})-\Phi(\boldsymbol{\tau}_i,\boldsymbol{\tau}_{-i})\\
=&\Phi(\boldsymbol{\tau}_i^{\varepsilon}(T_f),\boldsymbol{\tau}_{-i}(T_f))-\Phi(\boldsymbol{\tau}_i(T_f),\boldsymbol{\tau}_{-i}(T_f))
\end{align*}
Divide both sides by $\varepsilon>0$, and:
\begin{align*}
&\frac{{J}_i(\pi_{\boldsymbol{\theta}_i+\varepsilon\boldsymbol{v}_i},\pi_{-i})-{J}_i(\pi_{\boldsymbol{\theta}_i},\pi_{-i})}{\varepsilon}\\
=&\frac{\Phi(\boldsymbol{\tau}_i^{\varepsilon}(T_f),\boldsymbol{\tau}_{-i}(T_f))-\Phi(\boldsymbol{\tau}_i(T_f),\boldsymbol{\tau}_{-i}(T_f))}{\varepsilon}
\end{align*}
Taking the limit as $\varepsilon\to 0$ shows that Eq. \eqref{eq:grad_potential} holds.
\end{proof}
}

\subsection{Driver Heterogeneity Integration}
As stated in the definition of the potential function, all vehicles share a common potential, which limits the model's ability to capture diverse driving styles. To address this, we extend the exact potential function to a weighted potential function \cite{candogan2011flows} by \review{introducing a learnable weight $w_i>0$ to characterize the relative importance of the safety cost in the total cost for vehicle $i$. Then we redefine their stage cost as 
\begin{align}
c_i(\mathbf{s}(t),\mathbf{a}(t))
&\triangleq c_{i,\text{ind}}(t) \;+\; w_ic_{i,\text{safety}}(t),
% \label{eq:ci_split}  
\end{align}
Now we can reconstruct a potential function $\Phi^w$ as
\begin{align}
\begin{aligned}
\Phi^w(\boldsymbol{\tau}(T_f))\triangleq &\sum_{i=1}^{N}\frac{\phi_i(\mathbf{s}_i(T_f))}{w_i}
+ \sum_{t=1}^{T_f}\Bigg[\sum_{i=1}^{N} \frac{c_{i,\text{ind}}(t)}{w_i}\\
&+ \lambda_{\text{safety}}\sum_{i<j} f_{ij}(t)\Bigg].
\end{aligned}
\end{align}
As a result, we show the game can be viewed as a weighted potential game as follows: 
\begin{theorem} Under the conditions of Theorem 1, with the introduction of the sensitivity weights $w_i (i=1,\cdots, N)$, we have the following property:
\begin{align}
\begin{aligned}
     &{J}_i(\pi_i,\pi_{-i}) - {J}_i(\pi_i',\pi_{-i}) \\
     =& w_i[\Phi^w(\boldsymbol{\tau}_i(T_f),\boldsymbol{\tau}_{-i}(T_f)) - \Phi^w(\tilde{\boldsymbol{\tau}}_i(T_f),\boldsymbol{\tau}_{-i}(T_f))]
\end{aligned}
\end{align}
\label{thm2}
\end{theorem}
\begin{proof}
With the permutation invariance, it is easy to see:
 \begin{align*}
&w_i[\Phi^w(\boldsymbol{\tau}_i(T_f),\boldsymbol{\tau}_{-i}(T_f)) - \Phi^w(\tilde{\boldsymbol{\tau}}_i(T_f),\boldsymbol{\tau}_{-i}(T_f))]=\\
=&\Bigl[\phi_i(\mathbf{s}_i(T_f))-\phi_i(\tilde{\mathbf{s}}_i(T_f))\Bigr]+\sum_{t=1}^{T_f}\Bigl[\Bigl(c_{i,\text{ind}}(t) - \tilde{c}_{i,\text{ind}}(t)\Bigr)\\
&+w_i\Bigl(c_{i,\text{safety}}(t) - \tilde{c}_{i,\text{safety}}(t)\Bigr)\Bigr]\\
=&\Bigl[\phi_i(\mathbf{s}(T_f))-\phi_i(\tilde{\mathbf{s}}(T_f))\Bigr]\\
&+\sum_{t=1}^{T_f}\Bigl(c_i(\boldsymbol{s}(t),\mathbf{a}(t))-c_i(\tilde{\boldsymbol{s}}(t),\tilde{\mathbf{a}}(t))\Bigr)\\
=&J_i(\pi_i,\pi_{-i})-J_i(\tilde{\pi}_i,\pi_{-i})
 \end{align*}
which completes the proof.
\end{proof}
}

\review{Here, the weight $w_i>0$ can be interpreted as the sensitivity of vehicle $i$ to the safety cost: a larger $w_i$ indicates a more conservative vehicle, as it is more responsive to safety cost changes, whereas a smaller $w_i$ reflects a more aggressive driving style. In this way, we preserve the desirable properties of potential games while simultaneously incorporating individual driving styles. It is important to note that the weight is positive to ensure the gradient for each driver aligns in the same direction for the consistency of the optimization dynamics.}

\subsection{Deep Policy Learning}
\begin{figure*}[htbp]
  \centering
  \includegraphics[width=.85\linewidth]{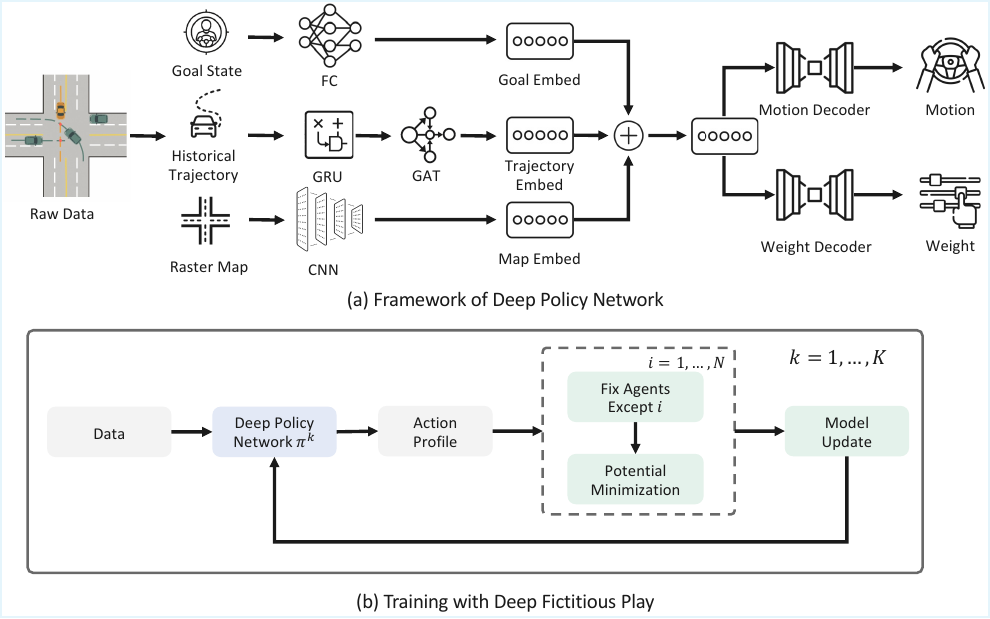}
  \caption{Deep policy network and training frameworks. (a) The deep policy network consists of a raster map encoder, a historical trajectory encoder, a goal state encoder, and two decoders—one for motion generation and the other for predicting driving style weights. (b) The training framework of Deep Fictitious Play with a theoretical guarantee.}
  \label{fig:framework}
\end{figure*}

To achieve human-like driving, the most efficient way is to learn from naturalistic driving datasets. Specifically, we propose a deep policy network $\pi_\theta$ parameterized with $\theta$ that incorporates a semantic encoder, a motion decoder, and a weight decoder as shown in Fig. \ref{fig:framework} (a). 

First, we convert the intersection map into a rasterized representation, where road boundaries are assigned a value of 1, the historical trajectory of the ego vehicle is assigned 0.5, and those of other vehicles are assigned 0.1. Afterwards, a Convolutional Neural Network (CNN) is employed followed by a fully connected network (FC) to extract general spatial features from the scene:
\begin{align}
    \mathbf{z}^{\text{map}} = \mathrm{FC}\left( \mathrm{CNN}(\mathbf{M})\right)\in \mathbb{R}^d
\end{align}
where $\mathbf{M}$ is the augmented map mentioned above. 

Second, to further capture interactions among vehicles, we employ a Gated Recurrent Unit (GRU) and a Graph Attention Network (GAT) to process the complex spatiotemporal patterns in historical trajectories. In detail, we first use a GRU to encode the historical trajectory of each vehicle into an embedding. These embeddings are then used to construct a fully connected graph $\mathcal{G}$, upon which a GAT is applied to learn the dependencies among vehicles:
\begin{align}
    &\mathbf{z}_i^{\text{GRU}} = \mathrm{GRU}(\mathbf{h}_i)\in \mathbb{R}^d\\
    &\alpha_{ij} = \frac{
\exp\left(\mathrm{LeakyReLU}\left( \mathbf{v}^\top [\mathbf{W} \mathbf{z}_i \, \| \, \mathbf{W} \mathbf{z}_j] \right)\right)
}{
\sum_{k \in \mathcal{N}} \exp\left(\mathrm{LeakyReLU}\left( \mathbf{v}^\top [\mathbf{W} \mathbf{z}_i \, \| \, \mathbf{W} \mathbf{z}_k] \right)\right)
}\\
&\mathbf{z}_i^{\mathrm{GAT}} = \mathrm{ReLU} \left( 
\sum_{j \in \mathcal{N}} \alpha_{ij} \cdot \mathbf{W} \mathbf{z}_j
\right)
\end{align}
where $\mathbf{v}$ and $\mathbf{W}$ are learnable parameters; $\|$ represents a concatenation operator.

Last, the goal embedding of the ego vehicle $i$ is simply derived from another shared fully connected network FC$_g$:
\begin{align}
    \mathbf{z}^{\text{goal}} = \mathrm{FC}_g(\mathbf{g}_i)
\end{align}

After obtaining the semantic embeddings, we employ two additional fully connected networks as the motion and weight decoder to generate the predicted motions:
\begin{align}
    \hat{\mathbf{a}}_i = \mathrm{FC}_1(\mathbf{z}^{\text{map}}\|\mathbf{z}^{\text{GAT}}_i\|\mathbf{z}^{\text{goal}})\\
    w_i = \text{sp}(\mathrm{FC}_2(\mathbf{z}^{\text{map}}\|\mathbf{z}^{\text{GAT}}_i\|\mathbf{z}^{\text{goal}}))
\end{align}

\subsection{Training with Deep Fictitious Play}
To fully leverage the capacity of deep learning models while maintaining theoretical guarantees, we train the deep policy using the Deep Fictitious Play (DFP) framework. Specifically, after the deep policy generates the motions, DFP further optimizes the potential function of one agent at a time while keeping the policies of all other agents fixed during each iteration, as shown in Algorithm 1 and Fig. \ref{fig:framework} (b). 

\review{
\begin{algorithm}[H]
\caption{Deep Fictitious Play (DFP)}
\label{alg:DFP}
\begin{algorithmic}[1]
\State Initialize the parameters $\boldsymbol{\theta}_i^{k}$ of policy $\pi_{\boldsymbol{\theta}_i}$ for each vehicle $i$ at iteration $k$.
\For{$i=1$ \textbf{to} $N$} \Comment{Initialize the actions.}
\State $\mathbf{a}_i^k=\pi_{\boldsymbol{\theta}^k_i}(\mathbf{s}_i(t))$
\EndFor
\For{$i=1$ \textbf{to} $N$} \Comment{Sequential update.}
    \State $\displaystyle \mathbf{a}_i^{k+1} \gets \arg\min_{\mathbf{a}_i \in \mathcal{A}}  J^{(i)}\bigl(\mathbf{s}_i(0),\mathbf{a}_i;\mathbf{a}^{k+1}_{<i},\mathbf{a}^k_{>i}\bigr)$     
    \State Others stay fixed.
    \EndFor
\State Update $\boldsymbol{\theta}_i^{k+1}$ from $\boldsymbol{\theta}_i^{k}$ with $\mathbf{a}_i=\pi_{\boldsymbol{\theta}_i}(\mathbf{s}_i(t))$ using gradient descent.
\end{algorithmic}
\end{algorithm}}

\review{\subsubsection{Alternating Best--Response Dynamics} The DFP framework adopts an iterative process to update vehicle actions. In each iteration $k$, each vehicle $i$ updates its own action $\mathbf{a}_i^{k}$  by minimizing its individual cost $J_i$ while holding other vehicles’ actions fixed. This update is repeated for \(K\) full passes over all vehicles before the resulting action profile is adopted. }

\review{\subsubsection{Network Loss Function} The resulting action profile can be used to generate each vehicle's trajectory, which then guides the training of the policy network. We use the Root Mean Squared Error (RMSE) between generated and reference trajectories for each vehicle as the final loss function}:
\begin{align}
    &\text{RMSE} = \sqrt{\frac{1}{T_f}\sum_{t=1}^{T_f}[(x_t-\hat{x}_t)^2+(y_t-\hat{y}_t)^2]}
\end{align}
where ${x}_t$ and ${y}_t$ are ground-truth values.

\subsubsection{Equilibrium Properties} Although the training process appears straightforward, the integration of Deep Fictitious Play with a potential differential game framework provides theoretical Nash equilibrium guarantees. \review{Before we proceed into the convergence proof, we introduce the following basic assumptions to ensure that the problem is well-defined:

\begin{assumption} All the vehicles have feasible actions at any time $t$, i.e., $\mathcal{A}$ is always non-empty and compact.
\label{ass:ABR}
\end{assumption}

\begin{assumption}  
The value of the potential function $\Phi^w$ has a finite lower bound, i.e., there exists $\underline{\Phi}>-\infty$ such that for any feasible trajectory $\boldsymbol{\tau}$, we have
    $\Phi^w(\boldsymbol{\tau})\geq \underline{\Phi}$.
\label{ass:compact_GS}
\end{assumption}

Here, we make no convexity assumptions, though they may yield stronger classical results. Under Assumption~\ref{ass:ABR} and cost component continuity, we obtain the following lemma:
\begin{lemma}
At each iteration $k$ in Algorithm~\ref{alg:DFP}, there is at least one feasible solution to each vehicle's subproblem.
\label{lem:existBR_GS}
\end{lemma}
\noindent To see this, since the cost components are continuous, $J_i$ is also continuous. Hence, under Assumption~\ref{ass:ABR}, the result follows directly from the Weierstrass extreme value theorem. 

Given this feasibility, we further show the potential function is monotonically nonincreasing as Algorithm~\ref{alg:DFP} progresses.
}

\review{\begin{lemma}[One-step Descent]
Fix the iteration $k$ and initial states $\mathbf{s}(0)$. Suppose $\mathbf a^{k}=(\mathbf a_i^{k},\mathbf a_{-i}^{k})$ denote action profile induced by the individual policies $\pi_{\boldsymbol{\theta}_1^k},\cdots,\pi_{\boldsymbol{\theta}_N^k}$. Let $\boldsymbol{\tau}^k$ denote the set of trajectories of all the $N$ vehicles generated from a fixed initial state $\mathbf{s}(0)$ under the action profile $\mathbf a^{k}$. Then, the updating pass in Algorithm~\ref{alg:DFP} will not increase the value of the potential function, i.e., $\Phi^w(\boldsymbol{\tau}^{k+1})\le\Phi^w(\boldsymbol{\tau}^{k})$.
\label{lem:descent_GS}
\end{lemma}
\begin{proof}
By Theorem \ref{thm2} and the definition in Eq. \eqref{Jdef}, we have
\begin{align}
\begin{aligned}
&J^{(i)}(\boldsymbol{\tau}_i^{k+1};\boldsymbol{\tau}_{<i}^{k+1},\boldsymbol{\tau}_{>i}^{k})
-
J_i(\boldsymbol{\tau}_i^{k};\boldsymbol{\tau}_{<i}^{k+1},\boldsymbol{\tau}_{>i}^{k})\\
=&
w_i[\Phi^w(\boldsymbol{\tau}_i^{k+1};\boldsymbol{\tau}_{<i}^{k+1},\boldsymbol{\tau}_{>i}^{k})-\Phi^w(\boldsymbol{\tau}_i^{k};\boldsymbol{\tau}_{<i}^{k+1},\boldsymbol{\tau}_{>i}^{k})],
\end{aligned}
\label{eqthm2}
\end{align}
with $w_i>0$. Note that when the initial states $\mathbf{s}(0)$ are fixed, the differences between the induced trajectories $\boldsymbol{\tau}_i^{k+1}$ and $\boldsymbol{\tau}_i^{k}$ of the vehicle $i$ are uniquely determined by the actions they take $\mathbf{a}_i^{k+1}$ and $\mathbf{a}_i^{k}$. Due to the exact minimization of $J^{(i)}$ by $\mathbf{a}_i^{k+1}$ in the update step of Algorithm~\ref{alg:DFP}, we have the inequality:
\begin{align*}
&J^{(i)}(\boldsymbol{\tau}_{<i}^{k+1}, \boldsymbol{\tau}_i^{k+1},\boldsymbol{\tau}_{>i}^{k})=J^{(i)}\bigl(\mathbf{s}_i(0),\mathbf{a}_i^{k+1};\boldsymbol{\tau}_{<i}^{k+1}, \boldsymbol{\tau}_{>i}^{k}\bigr)\\
\leq   &J^{(i)}\bigl(\mathbf{s}_i(0),\mathbf{a}_i^{k};\boldsymbol{\tau}_{<i}^{k+1}, \boldsymbol{\tau}_{>i}^{k}\bigr)=J^{(i)}(\boldsymbol{\tau}_{<i}^{k+1}, \boldsymbol{\tau}_i^{k},\boldsymbol{\tau}_{>i}^{k})
% &=J^{(i)}(\boldsymbol{\tau}_{<{i-1}}^{k+1}, \boldsymbol{\tau}_{i-1}^{k+1},\boldsymbol{\tau}_{>{i-1}}^{k})
\end{align*}
By Eq. \eqref{eqthm2} and $w_i>0$, we can obtain
\begin{align*}
\Phi^w(\boldsymbol{\tau}_i^{k+1};\boldsymbol{\tau}_{<i}^{k+1},\boldsymbol{\tau}_{>i}^{k})&\leq \Phi^w(\boldsymbol{\tau}_i^{k};\boldsymbol{\tau}_{<i}^{k+1},\boldsymbol{\tau}_{>i}^{k})\\
&=\Phi^w(\boldsymbol{\tau}_{i-1}^{k+1};\boldsymbol{\tau}_{<{i-1}}^{k+1},\boldsymbol{\tau}_{>{i-1}}^{k})
\end{align*}
We repeatedly use this inequality by letting $i$ run through $1$ to $N$, and then we can obtain:
\begin{align*}
\Phi^w(\boldsymbol{\tau}^{k+1})&=\Phi^w(\boldsymbol{\tau}_N^{k+1};\boldsymbol{\tau}_{<N}^{k+1},\boldsymbol{\tau}_{>N}^{k})\leq \cdots\\
&\leq \Phi^w(\boldsymbol{\tau}_1^{k+1};\boldsymbol{\tau}_{<1}^{k+1},\boldsymbol{\tau}_{>1}^{k})=\Phi^w(\boldsymbol{\tau}^{k}).
\end{align*}
\end{proof}}

\review{With Lemma \ref{lem:descent_GS}, we see the sequence $\{\Phi^w(\boldsymbol{\tau}^{k})\}$ is monotonically non‑increasing as $k$ increases. Under Assumption~\ref{ass:compact_GS}, it is lower bounded. Thus, by the monotone convergence theorem, it is easy to see the existence of its limit:}

\begin{corollary}[Existence of the Limit]\label{lem:limit_GS}
The sequence $\{\Phi^w(\boldsymbol{\tau}^{k})\}$ converges to its infimum, i.e.,
$\displaystyle\lim_{k\to\infty}\Phi^w(\boldsymbol{\tau}^{k})=\Phi^{\infty}$, where $\Phi^{\infty}=\inf_{k\in\mathbb{Z}^+}{\Phi^w(\boldsymbol{\tau}^{k})}$ is finite. 
\label{coro:limit}
\end{corollary}

\review{Now we present our theoretical result for DFP, which relies on the following regularity condition:
\begin{assumption}
For each vehicle $i$ and for any fixed feasible action profile of other vehicles, the best-response subproblem in Algorithm~\ref{alg:DFP} has a unique minimizer.
\label{assmp:uniqueness}
\end{assumption}

Under this assumption, the following theorem establishes the equilibrium convergence properties of DFP.

\begin{theorem}[Nash Equilibrium]
Suppose $\{\mathbf a^{k}\}_{k=1,2,\cdots}$ is the action sequence generated during the iterative process in Algorithm~\ref{alg:DFP}. Let $\mathbf{a}^{\infty}$ denote a cluster point of the sequence such that for $\forall \delta>0$, there exists $k$ satisfying $0<||\mathbf{a}^{k}-\mathbf{a}^{\infty}||<\delta$. Then, $\mathbf{a}^{\infty}$ is a Nash equilibrium of the weighted potential differential game. If it is a unique cluster point, then $\{\mathbf a^{k}\}_{k=1,2,\cdots}$ converges to $\mathbf a^{\infty}$.
\label{lem:clusterNE_GS}
\end{theorem}

\begin{proof}
Given a cluster point $\mathbf{a}^{\infty}$ of the action sequence $\{\mathbf a^{k}\}_{k=1,2,\cdots}$, by the properties of cluster points, we know there exists a subsequence of the action sequence $\{\mathbf a^{k_m}\}_{m=1,2,\cdots}$ where $k_1<k_2<\cdots$ such that $\mathbf{a}^{k_m}\to\mathbf{a}^{\infty}$. Since $\mathcal{A}$ is compact, so it is closed, which indicates $\mathbf{a}^{\infty}\in \mathcal{A}$. Let $\boldsymbol{\tau}^{\infty}$ be its induced trajectory. Thus, it is a possible solution to the game. With the initial states fixed, this gives rise to a sequence of corresponding trajectories $\{\boldsymbol{\tau}^{k_m}\}_{m=1,2,\cdots}$ with a limit of $\boldsymbol{\tau}^{\infty}$. Meanwhile, under Assumption~\ref{assmp:uniqueness}, by Berge's maximum theorem, we can build a continuous map $T$ between $\mathbf{a}^{k}$ and $\mathbf{a}^{k+1}$ as $\mathbf{a}^{k+1}=T(\mathbf{a}^{k})$, which leads to $\mathbf{a}^{k_m+1}=T(\mathbf{a}^{k_m})\to T(\mathbf{a}^{\infty})\in\mathcal{A}$. Let $\hat{\mathbf{a}}=T(\mathbf{a}^{\infty})$. If $\hat{\mathbf{a}}\neq \mathbf{a}^{\infty}$, by Lemma~\ref{lem:descent_GS}, its induced trajectory $\hat{\boldsymbol{\tau}}$ should satisfy $\Phi^w(\hat{\boldsymbol{\tau}})< \Phi^w(\boldsymbol{\tau}^{\infty})$. But Corollary~\ref{coro:limit} gives $\Phi^w(\hat{\boldsymbol{\tau}})=\Phi^{\infty}$. This contradiction suggests $\hat{\mathbf{a}}=T(\mathbf{a}^{\infty})=\mathbf{a}^{\infty}$. Hence, we have $\mathbf{a}^{k_m+1}\to\mathbf{a}^{\infty}$ as well.

Now we show the convergence of DFP to the Nash equilibrium. (i) First, we show any cluster point $\mathbf{a}^{\infty}$ is a Nash equilibrium by contradiction as well. Assume it is not a Nash equilibrium. There exists a unilateral $\tilde{\mathbf{a}}_i$ inducing a trajectory $\tilde{\boldsymbol{\tau}}_i$ such that 
\begin{align*}
J^{(i)}(\tilde{\boldsymbol{\tau}}_i,\boldsymbol{\tau}^{\infty}_{-i})<J^{(i)}(\boldsymbol{\tau}^{\infty}_i,\boldsymbol{\tau}^{\infty}_{-i})    
\end{align*} 
Let $\eta \triangleq J^{(i)}(\boldsymbol{\tau}^{\infty}_i,\boldsymbol{\tau}^{\infty}_{-i})-J^{(i)}(\tilde{\boldsymbol{\tau}}_i,\boldsymbol{\tau}^{\infty}_{-i})>0$. Since $\mathbf{a}^{k_m}\to\mathbf{a}^{\infty}$ and $\mathbf{a}^{k_m+1}\to\mathbf{a}^{\infty}$, for $\forall \varepsilon>0$, we can find a sufficiently large $M>0$ such that $||\mathbf{a}^{k_m}-\mathbf{a}^{\infty}||<\varepsilon$ and $||\mathbf{a}^{k_m+1}-\mathbf{a}^{\infty}||<\varepsilon$ for $\forall m>M$. As a result, for $\forall i$, we can also have $||\mathbf{a}_i^{k_m}-\mathbf{a}_i^{\infty}||<\varepsilon$, and similarly, $||\mathbf{a}_{-i}^{k_m}-\mathbf{a}_{-i}^{\infty}||<\varepsilon$. This will also ensure the similarity between the corresponding trajectories. By continuity of the cost components, $J^{(i)}$ is also continuous. Thus, we can always find such a $\varepsilon$ to restrict the difference between the actions, so that 
\begin{align*}
J^{(i)}(\tilde{\boldsymbol{\tau}}_i,\boldsymbol{\tau}^{k_{m}+1}_{<i},\boldsymbol{\tau}^{k_{m}}_{>i}) \leq J^{(i)}(\tilde{\boldsymbol{\tau}}_i, \boldsymbol{\tau}^{\infty}_{-i})+\frac{\eta}{4}    
\end{align*}
and
\begin{align*}
J^{(i)}(\boldsymbol{\tau}^{\infty}_i,\boldsymbol{\tau}^{\infty}_{-i}) \leq J^{(i)}(\boldsymbol{\tau}^{k_m}_i,\boldsymbol{\tau}^{k_{m}+1}_{<i}, \boldsymbol{\tau}^{k_m}_{>i})+\frac{\eta}{4}    
\end{align*}
which give
\begin{align*}
J^{(i)}(\tilde{\boldsymbol{\tau}}_i,\boldsymbol{\tau}^{k_{m}+1}_{<i},\boldsymbol{\tau}^{k_{m}}_{>i})\leq  J^{(i)}(\boldsymbol{\tau}^{k_m}_i,\boldsymbol{\tau}^{k_{m}+1}_{<i}, \boldsymbol{\tau}^{k_m}_{>i})-\frac{\eta}{2}
\end{align*}
Using Theorem~\ref{thm2} and the non-increasing nature of the updating steps, we have
\begin{align*}
\Phi^{w}(\tilde{\boldsymbol{\tau}}_i,\boldsymbol{\tau}^{k_{m}+1}_{<i},\boldsymbol{\tau}^{k_{m}}_{>i})&\leq  \Phi^{w}(\boldsymbol{\tau}^{k_m}_i,\boldsymbol{\tau}^{k_{m}+1}_{<i}, \boldsymbol{\tau}^{k_m}_{>i})-\frac{\eta}{2w_i}
\end{align*}
Noting that
\begin{align*}
\Phi^{w}(\boldsymbol{\tau}^{k_m}_i,\boldsymbol{\tau}^{k_{m}+1}_{<i}, \boldsymbol{\tau}^{k_m}_{>i})\leq \Phi^{w}(\boldsymbol{\tau}^{k_m}_{i-1},\boldsymbol{\tau}^{k_{m}+1}_{<i-1}, \boldsymbol{\tau}^{k_m}_{>i-1}),
\end{align*}
we can let $i$ run down to 1 and repeatedly use the inequality, which results in
\begin{align}
\begin{aligned}
\Phi^{w}(\tilde{\boldsymbol{\tau}}_i,\boldsymbol{\tau}^{k_{m}+1}_{<i},\boldsymbol{\tau}^{k_{m}}_{>i})\leq \Phi^{w}(\boldsymbol{\tau}^{k_m})-\frac{\eta}{2w_i}
\end{aligned}
\label{rhsineq}
\end{align}
Meanwhile, by the optimality of $\mathbf{a}^{k_{m}+1}_i$, we have
\begin{align*}
&J^{(i)}(\tilde{\boldsymbol{\tau}}_i,\boldsymbol{\tau}^{k_{m}+1}_{<i},\boldsymbol{\tau}^{k_{m}}_{>i})\geq J^{(i)}(\boldsymbol{\tau}^{k_{m}+1}_i,\boldsymbol{\tau}^{k_{m}+1}_{<i},\boldsymbol{\tau}^{k_{m}}_{>i})  
\end{align*}
Using Theorem~\ref{thm2} and applying the optimality condition, we have
\begin{align*}
\Phi^w(\tilde{\boldsymbol{\tau}}_i,\boldsymbol{\tau}^{k_{m}+1}_{<i},\boldsymbol{\tau}^{k_{m}}_{>i})&\geq \Phi^w(\boldsymbol{\tau}^{k_{m}+1}_i,\boldsymbol{\tau}^{k_{m}+1}_{<i},\boldsymbol{\tau}^{k_{m}}_{>i})\\
&\geq \Phi^{w}(\boldsymbol{\tau}^{k_{m}+1}_{i+1},\boldsymbol{\tau}^{k_{m}+1}_{<i+1},\boldsymbol{\tau}^{k_{m}}_{>i+1})\geq \cdots\\
&\geq \Phi^{w}(\boldsymbol{\tau}^{k_{m}+1}).   
\end{align*}
With Eq. \eqref{rhsineq}, we see
\begin{align*}
\Phi^{w}(\boldsymbol{\tau}^{k_{m}+1})\leq \Phi^w(\boldsymbol{\tau}^{k_m})-\frac{\eta}{2w_i}
\end{align*}
which contradicts the convergence of $\Phi^w$. Therefore, $\mathbf{a}^{\infty}$ must be a Nash equilibrium. (ii) Now we show if $\mathbf{a}^{\infty}$ is a unique cluster point, then the action sequence $\{\mathbf{a}^k\}_{k=1,2,\cdots,}$ converges to the Nash equilibrium. We still prove by contradiction. Suppose $\{\mathbf{a}^k\}_{k=1,2,\cdots}$ does not converge to $\mathbf{a}^{\infty}$. Then there exists some $\delta>0$ and an infinite subsequence $\{\mathbf{a}^{l_m}\}_{m=1,2,\cdots}$ such that $\forall m=1,2,\cdots,||\mathbf{a}^{l_m}-\mathbf{a}^{\infty}||\geq \delta$. By Bolzano-Weierstrass theorem, we have a convergent subsequence of $\{\mathbf{a}^{l_m}\}_{m=1,2,\cdots}$, denoted by $\{\mathbf{a}^{l_{m_j}}\}_{j=1,2,\cdots}$. Denote the limit as $\mathbf{a}_l^{\infty}\in\mathcal{A}$. So it is easy to see $||\mathbf{a}_l^{\infty}-\mathbf{a}^{\infty}||\geq \delta$, which indicates $\mathbf{a}_l^{\infty}\neq \mathbf{a}^{\infty}$. This gives a different cluster point, and thus contradicts to the uniqueness of $\mathbf{a}^{\infty}$. As a result, the action sequence $\{\mathbf{a}^k\}_{k=1,2,\cdots}$ must converge to $\mathbf{a}^{\infty}$.

\end{proof}}

\review{In practical settings, the exact best response may not be obtained at every update step due to real-world uncertainties or imperfect information. To account for this limitation, we consider an approximate best-response setting instead.} During each updating iteration $k$, when the vehicle $i$ is updated with $\mathbf a_{-i}$ fixed, let $\widehat{\mathbf a}_i^{k}$ denote the approximate solution to the exact best response problem which has the exact solution $\mathbf a_i^{k}\in \mathop{\arg\min}_{\mathbf a_i\in\mathcal A} J^{(i)}(\mathbf{s}_i(0), \mathbf a_i;\hat{\mathbf a}_{<i}^{k},\hat{\mathbf a}_{>i}^{k-1})$. We define the sub-optimality gap as $\varepsilon_i^{k}=J^{(i)}(\mathbf{s}_i(0), \hat{\mathbf{a}}^k_i;\hat{\mathbf a}_{<i}^{k},\hat{\mathbf{a}}_{>i}^{k-1})-J^{(i)}(\mathbf{s}_i(0), \mathbf{a}^k_i;\hat{\mathbf a}_{<i}^{k},\hat{\mathbf a}_{>i}^{k-1})\ge0$ and denote the maximal error as $\Delta_k=\max_i \varepsilon_i^{k}$. We further impose the following assumption on the approximation errors:
\begin{assumption}\label{ass:ABR_error}
The total maximal approximation error is bounded, i.e., $\sum_{k=0}^{\infty}\Delta_k<\infty$.
\end{assumption}

Now we show the following property of DFP:

\review{
\begin{lemma}
Fix the iteration k and initial
states $\mathbf{s}(0)$. Assume all the updated actions are approximate actions, and the trajectory induced by the approximate action
$\widehat{\mathbf a}_i^{k}$ is denoted by  $\hat{\boldsymbol{\tau}}_i^k$. Let $w_{\min}=\min_i w_i>0$ and define $\widetilde\Delta_k=
\sum_{i=1}^{N}w_i^{-1}\varepsilon_i^{k}$. Then each iteration of updating in Algorithm~\ref{alg:DFP} yields
\begin{align*}
\Phi^w(\hat{\boldsymbol{\tau}}^{k})\leq \Phi^w(\hat{\boldsymbol{\tau}}^{k-1}) +\widetilde\Delta_k.
\end{align*}
\label{lem:perturb_GS}
\end{lemma}

\begin{proof}
Note that $\mathbf{a}_i^k=\mathop{\arg\min}_{\mathbf{a}_i}J^{(i)}(\mathbf{s}_i(0), \mathbf{a}_i;\hat{\mathbf a}_{<i}^{k},\hat{\mathbf a}_{>i}^{k-1})$. Thus, for any $\mathbf{a}_i\in\mathcal{A}$, we have 
\begin{align*}
J^{(i)}(\mathbf{s}_i(0), \mathbf{a}^k_i;\hat{\mathbf a}_{<i}^{k},\hat{\mathbf a}_{>i}^{k-1})\leq J^{(i)}(\mathbf{s}_i(0), \mathbf{a}_i;\hat{\mathbf a}_{<i}^{k},\hat{\mathbf a}_{>i}^{k-1})    
\end{align*}
which can be rewritten as
\begin{align*}
J_i(\boldsymbol{\tau}_i^{k},\hat{\boldsymbol{\tau}}_{<i}^{k},\hat{\boldsymbol{\tau}}_{>i}^{k-1})\leq 
J_i(\boldsymbol{\tau}_i,\hat{\boldsymbol{\tau}}^k_{<i},\hat{\boldsymbol{\tau}}_{>i}^{k-1}).
\end{align*}
where $\boldsymbol{\tau}_i$ is the trajectory induced by any feasible action $\mathbf{a}_i\in\mathcal{A}$. This results in the following inequality:
\begin{align*}
J_i(\hat{\boldsymbol{\tau}}^k_i,\hat{\boldsymbol{\tau}}_{<i}^{k},\hat{\boldsymbol{\tau}}_{>i}^{k-1})\leq 
J_i(\boldsymbol{\tau}_i,\hat{\boldsymbol{\tau}}_{<i}^k,\hat{\boldsymbol{\tau}}_{>i}^{k-1})
+\varepsilon_i^{k}.   
\end{align*}
Let $\boldsymbol{\tau}_i=\hat{\boldsymbol{\tau}}_i^{k-1}$. We can easily obtain 
\begin{align*}
J_i(\hat{\boldsymbol{\tau}}^k_i,\hat{\boldsymbol{\tau}}_{<i}^{k},\hat{\boldsymbol{\tau}}_{>i}^{k-1})\leq 
J_i(\hat{\boldsymbol{\tau}}_i^{k-1},\hat{\boldsymbol{\tau}}_{<i}^{k},\hat{\boldsymbol{\tau}}_{>i}^{k-1})
+\varepsilon_i^{k}
\end{align*}
By Theorem \eqref{thm2} and $w_i>0$, we have
\begin{align*}
&J_i(\hat{\boldsymbol{\tau}}^k_i,\hat{\boldsymbol{\tau}}_{<i}^{k},\hat{\boldsymbol{\tau}}_{>i}^{k-1})-J_i(\hat{\boldsymbol{\tau}}^{k-1}_i,\hat{\boldsymbol{\tau}}_{<i}^{k},\hat{\boldsymbol{\tau}}_{>i}^{k-1})\\
=&w_i[\Phi^w(\hat{\boldsymbol{\tau}}^k_i,\hat{\boldsymbol{\tau}}_{<i}^{k},\hat{\boldsymbol{\tau}}_{>i}^{k-1})-\Phi^w(\hat{\boldsymbol{\tau}}^{k-1}_i,\hat{\boldsymbol{\tau}}_{<i}^{k},\hat{\boldsymbol{\tau}}_{>i}^{k-1})],
\end{align*}
Hence, we can obtain
\begin{align*}
\Phi^w(\hat{\boldsymbol{\tau}}^k_i,\hat{\boldsymbol{\tau}}_{<i}^{k},\hat{\boldsymbol{\tau}}_{>i}^{k-1})&\leq \Phi^w(\hat{\boldsymbol{\tau}}^{k-1}_i,\hat{\boldsymbol{\tau}}_{<i}^{k},\hat{\boldsymbol{\tau}}_{>i}^{k-1})+\varepsilon_i^{k}/w_i
\end{align*}
This is equivalent to 
\begin{align*}
\Phi^w(\hat{\boldsymbol{\tau}}_{\leq i}^{k},\hat{\boldsymbol{\tau}}_{>i}^{k-1})&\leq \Phi^w(\hat{\boldsymbol{\tau}}_{<i}^{k},\hat{\boldsymbol{\tau}}_{\geq i}^{k-1})+\varepsilon_i^{k}/w_i
\end{align*}
Let $i$ run from $1$ to $N$ and sum them up. Then, we have
\begin{align*}
\sum_{i=1}^N\Phi^w(\hat{\boldsymbol{\tau}}_{\leq i}^{k},\hat{\boldsymbol{\tau}}_{>i}^{k-1})\leq \sum_{i=1}^N\Phi^w(\hat{\boldsymbol{\tau}}_{<i}^{k},\hat{\boldsymbol{\tau}}_{\geq i}^{k-1})+\tilde{\Delta}_k
\end{align*}
Noting that $\Phi^w(\hat{\boldsymbol{\tau}}_{<i}^{k},\hat{\boldsymbol{\tau}}_{\geq i}^{k-1})=\Phi^w(\hat{\boldsymbol{\tau}}_{\leq i-1}^{k},\hat{\boldsymbol{\tau}}_{>i-1}^{k-1})$, we can cancel the equal items on two sides, and then obtain
\begin{align*}
\Phi^w(\hat{\boldsymbol{\tau}}^{k})\leq \Phi^w(\hat{\boldsymbol{\tau}}^{k-1})+\tilde{\Delta}_k
\end{align*}
\end{proof}}

With Lemma \ref{lem:perturb_GS}, we can show the convergence of DFP under the bounded approximation errors:
\begin{theorem}[Convergence under Errors]\label{cor:ABR_error}
Under Assumption~\ref{ass:ABR}-\ref{ass:ABR_error}, every cluster point $\hat{\mathbf{a}}^{\infty}$ of $\{\hat{\mathbf{a}}^{k}\}_{k=1, 2,\cdots}$ reaches a Nash equilibrium of the weighted potential differential game. If the cluster point $\hat{\mathbf{a}}^{\infty}$ is unique, then the action sequence $\{\hat{\mathbf{a}}^{k}\}$ converges to the Nash equilibrium.
\end{theorem}

\begin{proof}
With Lemma~\ref{lem:perturb_GS}, we have $\Phi^w(\hat{\boldsymbol{\tau}}^{k})\le \Phi^w(\hat{\boldsymbol{\tau}}^{k-1})+\widetilde\Delta_k$. Meanwhile, we can see by Assumption~\ref{ass:ABR_error} that
\begin{align*}
\sum_{k=0}^{\infty} \widetilde\Delta_{k}=\sum_{k=0}^{\infty}\sum_{i=1}^Nw_i^{-1}\varepsilon_i^k\leq w_{\min}^{-1}N\sum_{k=0}^{\infty}\Delta_k<\infty   
\end{align*} 
This indicates $\widetilde\Delta_k\to 0$. Now we define the tail error $r_k=\sum_{l=k+1}^{\infty}\tilde{\Delta}_l\geq 0$ which is upper bounded and satisfies $\lim_{k\to\infty}r_k= 0$. It is obvious that $r_{k-1}=r_{k}+\tilde{\Delta}_k$. Let $\tilde{\Phi}^{w}_k=\Phi^{w}(\hat{\boldsymbol{\tau}}^k)+r_k$. Then by Lemma~\ref{lem:perturb_GS}, we can obtain
\begin{align*}
  \tilde{\Phi}^{w}_k\leq \Phi^{w}(\hat{\boldsymbol{\tau}}^{k-1})+\tilde{\Delta}_{k}+r_{k}= \Phi^{w}(\hat{\boldsymbol{\tau}}^{k-1})+r_{k-1}=\tilde{\Phi}^{w}_{k-1}
\end{align*}
Since $\Phi^{w}$ is lower bounded and $r_k\geq 0$, $\tilde{\Phi}^{w}_k$ is lower bounded. Thus, by the monotone convergence theorem, $\tilde{\Phi}^{w}_k$ converges, and we denote the limit as $\tilde{\Phi}^w_{\infty}$. Using $r_k\to 0$ and the algebraic limit theorem, we have $\Phi^{w}(\hat{\boldsymbol{\tau}}^{k})\to\tilde{\Phi}^w_{\infty}$. Now given the cluster point $\hat{\mathbf{a}}^\infty$, we can find a subsequence $\{\hat{\mathbf{a}}^{k_m}\}_{m=1,2,\cdots}$ such that $\hat{\mathbf{a}}^{k_m}\to\hat{\mathbf{a}}^\infty$. Let $\hat{\boldsymbol{\tau}}^{\infty}$ be the trajectory induced by $\hat{\mathbf{a}}^\infty$. Then it is easy to see $\lim_{m\to\infty}\Phi^{w}(\hat{\boldsymbol{\tau}}^{k_m})=\Phi^w(\hat{\boldsymbol{\tau}}^{\infty})=\tilde{\Phi}^w_{\infty}$. We further consider the action sequence after one pass $\{\hat{\mathbf{a}}^{k_m+1}\}_{m=1,2,\cdots}$. Since $\mathcal{A}^{N}$ is compact, there must exist a convergent subsequence $\{\hat{\mathbf{a}}^{k_{m_l}+1}\}_{l=1,2,\cdots}$ such that $\hat{\mathbf{a}}^{k_{m_l}+1}\to\hat{\mathbf{a}}^{b}$ where $\hat{\mathbf{a}}^{k_{m_l}}\to\hat{\mathbf{a}}^{\infty}$. The induced trajectory $\hat{\boldsymbol{\tau}}^{b}$ by $\hat{\mathbf{a}}^{b}$ should satisfy $\Phi^w(\hat{\boldsymbol{\tau}}^{b})=\tilde{\Phi}^w_{\infty}$. Now we show $\hat{\mathbf{a}}^{b}$ can be continuously mapped from $\hat{\mathbf{a}}^\infty$. Actually, we can obtain $J^{(i)}(\hat{\boldsymbol{\tau}}_i^{k_{m_\ell}+1},
\hat{\boldsymbol{\tau}}_{<i}^{k_{m_\ell}+1},
\hat{\boldsymbol{\tau}}_{>i}^{k_{m_\ell}})\leq J^{(i)}(\boldsymbol{\tau}_i, \hat{\boldsymbol{\tau}}_{<i}^{k_{m_\ell}+1},
\hat{\boldsymbol{\tau}}_{>i}^{k_{m_\ell}})+\varepsilon_{i}^{k_{m_l}+1}$ for any feasible $\boldsymbol{\tau}_i$ induced by corresponding actions. Since $J^{(i)}$ is continuous and $\varepsilon_{i}^{k}\to 0$, taking $l\to\infty$ yields $J^{(i)}(\hat{\boldsymbol{\tau}}_i^b,
\hat{\boldsymbol{\tau}}_{<i}^{b},
\hat{\boldsymbol{\tau}}_{>i}^{\infty})\leq J^{(i)}(\boldsymbol{\tau}_i, \hat{\boldsymbol{\tau}}_{<i}^{b},
\hat{\boldsymbol{\tau}}_{>i}^{\infty})$. Therefore, the action $\hat{\mathbf{a}}_i^b$ that induces $\hat{\boldsymbol{\tau}}_i^b$ is an exact best response with others' responses as $(\hat{\mathbf{a}}_{<i}^b, \hat{\mathbf{a}}_{>i}^\infty)$. Let $i$ run from $1$ to $N$, and we can see $\hat{\mathbf{a}}^b=(\hat{\mathbf{a}}^b_1,\cdots, \hat{\mathbf{a}}^b_N)$ is exactly the result of one full exact cyclic best-response pass initialized at $\hat{\mathbf{a}}^{\infty}$, which indicates we have a well-defined map $T$ satisfying $\hat{\mathbf{a}}^b=T(\hat{\mathbf{a}}^{\infty})$. Then, with the contradiction techniques in the proof of Theorem~\ref{lem:clusterNE_GS}, we can prove $\hat{\mathbf{a}}^b=\hat{\mathbf{a}}^{\infty}$, and further show: the cluster point $\hat{\mathbf{a}}^{\infty}$ of $\{\hat{\mathbf{a}}^k\}_{k=1,2,\cdots}$ is a Nash equilibrium, and if it is unique, the action sequence will converge to this singleton point. 
\end{proof}

\begin{remark}[Interpretation via Level-$K$ Reasoning]\label{rem:levelK}
The outer-iteration index in Algorithm~\ref{alg:DFP} also admits an intuitive behavioral interpretation.
Specifically, the sequential best-response updates can be viewed as a form of iterative bounded rationality that is closely related to cognitive-hierarchy (CH) and level-$K$ models in behavioral game theory \cite{chen2024combining,gill2016cognitive}.
If we regard the initial policy profile as a level-$0$ baseline, then one outer iteration produces a level-$1$ profile by responding to the current opponents' policies, repeating this procedure yields a level-$K$ profile after $K$ outer iterations.
Under this view, DFP provides a constructive way to parameterize and learn progressively deeper strategic reasoning, while the potential-game structure ensures that these iterative updates follow a well-behaved descent of a global objective.
We emphasize that this is an interpretation rather than a strict equivalence, since classical level-$K$ models may posit an exogenous distribution over lower-level types, whereas our updates are generated endogenously by the learned best responses.
\end{remark}

\section{Experiments}
\subsection{Dataset}
\begin{figure}[!t] % !t 按 IEEEtran 规范优先放到顶端
  \centering
  %------------ 子图 A -------------
  \subfloat[Scenario MA]{%
    \includegraphics[width=.9\linewidth]{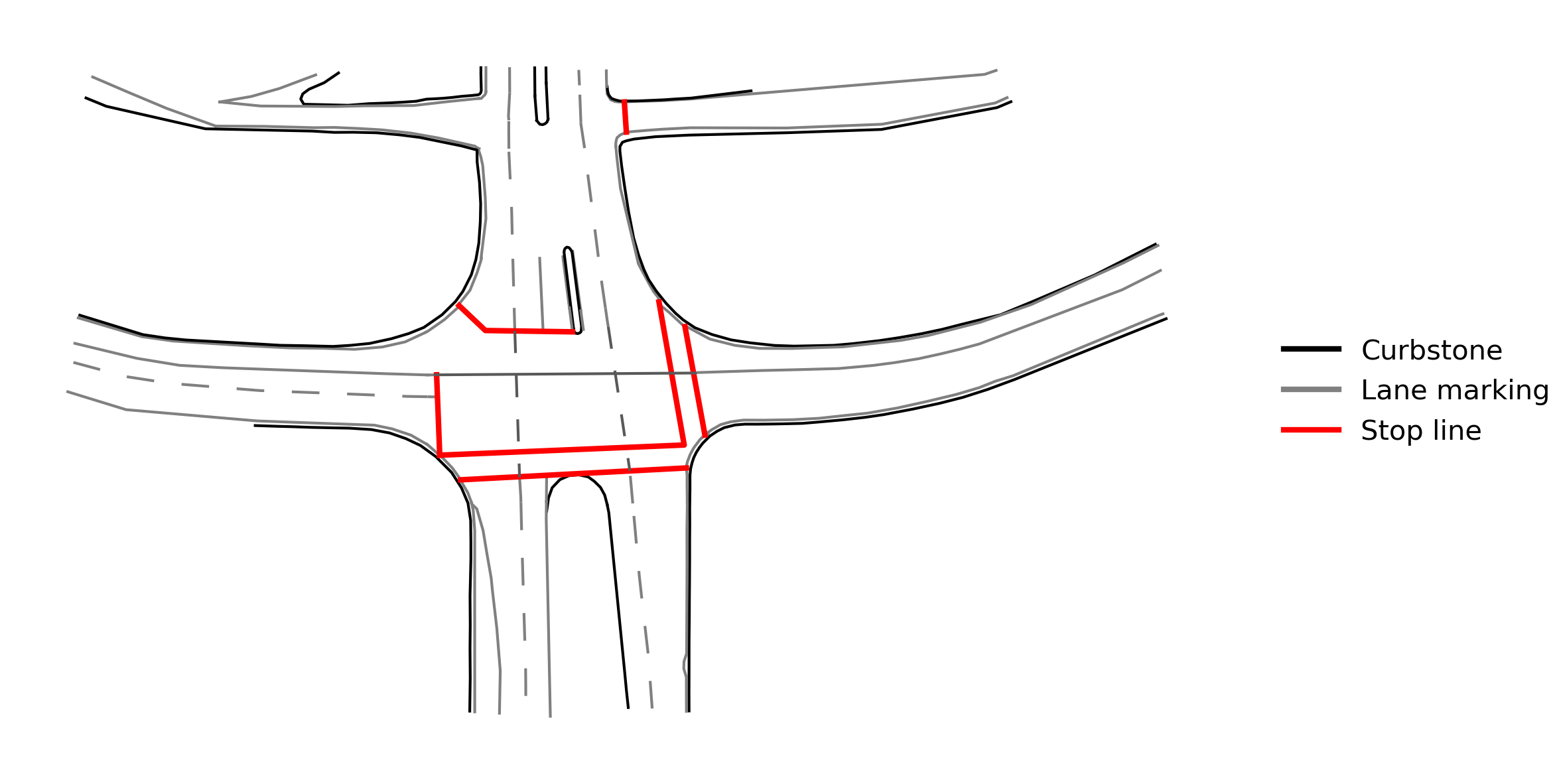}%
    \label{fig:ma}%
  }
  \hfill  % 或者 \hspace{0.02\linewidth} 手动控制间距
  %------------ 子图 B -------------
  \subfloat[Scenario GL]{%
    \includegraphics[width=.9\linewidth]{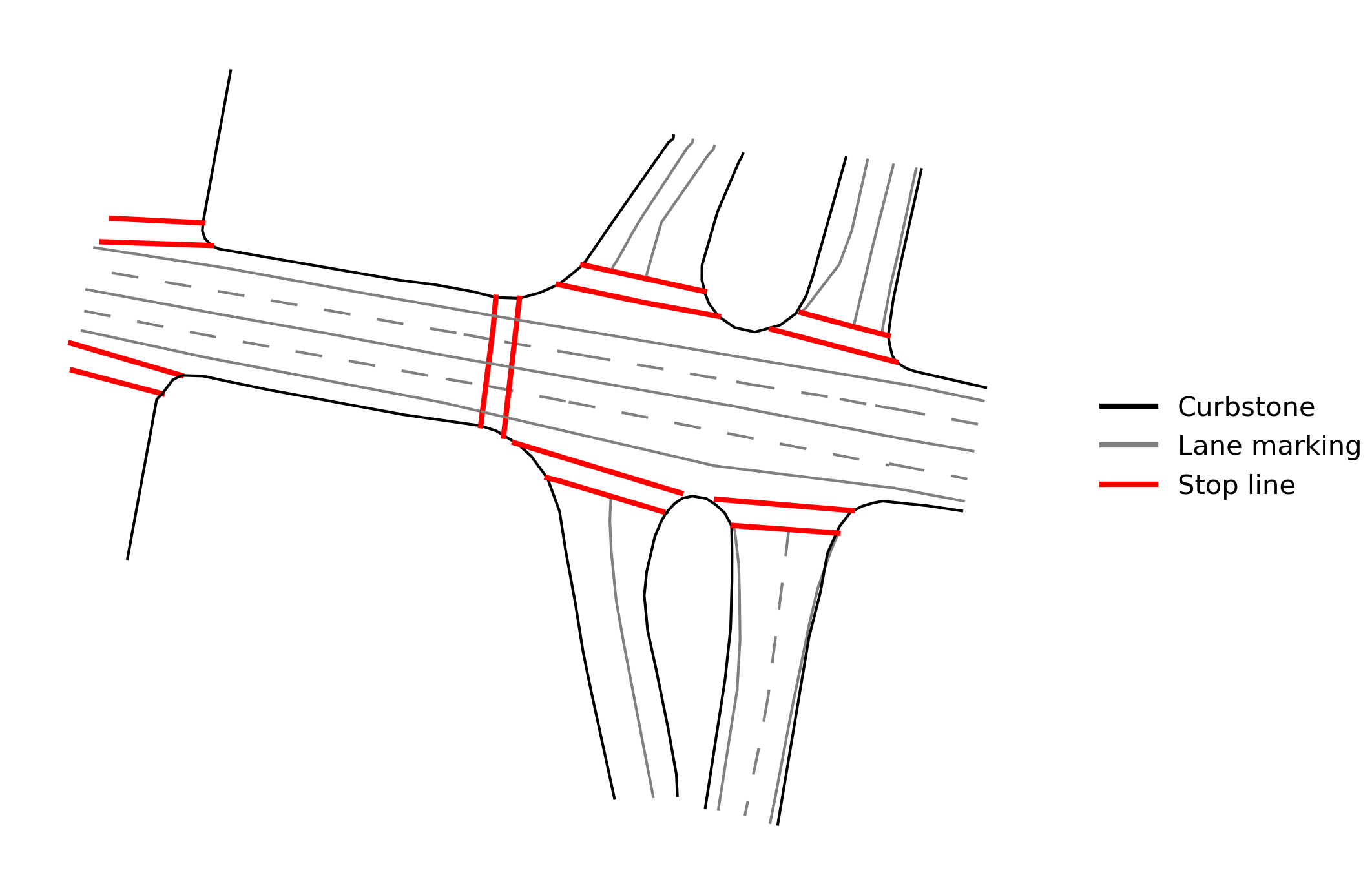}%
    \label{fig:gl}%
  }
  \caption{Visualization of MA and GL intersections.}
  \label{fig:scenarios}
\end{figure}

To evaluate the effectiveness of our framework, we conduct experiments using the INTERACTION dataset \cite{zhan2019interaction}, which features a variety of highly interactive driving scenarios, including merging, roundabouts, and unsignalized intersections. In this study, we select two U.S. scenarios—GL and MA—as illustrated in Fig.~\ref{fig:scenarios} for evaluation.

The raw data were first filtered to remain only vehicle trajectories with 40 consecutive frames within the designated spatial coverage area to ensure sufficient information of vehicles. To extract interacting vehicle pairs, we assigned a driving direction to each vehicle based on its position and velocity changes, and defined interaction based on conflicting movement directions as shown in Fig. \ref{fig:interacting}. For instance, a vehicle traveling northbound is considered to interact with vehicles coming from the eastbound and westbound directions (through or left turns), as well as those making left turns from the southbound direction. Upon analyzing the dataset, we observed that vehicles traveling straight were significantly more prevalent than turning vehicles. To mitigate this imbalance and ensure diverse interaction scenarios, we constructed each scene using a turning vehicle as the reference. All other trajectories interacting with this reference vehicle were then included in the scene. Using this approach, we generated 2,769 interaction scenes for the MA dataset and 1,938 scenes for the GL dataset.

To effectively evaluate the planning performance, we use the past one second of historical trajectories to plan the motion for the upcoming three second. Table \ref{tab:stats} presents the statistics of two scenarios. We split the dataset into 70\% for training and 30\% for testing.

\begin{table}[htbp]
  \centering
  \caption{Summary Statistics of the MA and GL Scenarios}
  \label{tab:stats}
  \renewcommand\arraystretch{1.2}
  \begin{tabular}{lcc}
    \hline
    \textbf{Metric} & \textbf{MA} & \textbf{GL} \\
    \hline
    Number of scenes                &    2769        &      1938  \\
    Avg.\ agents per scene          &   3.55         &      3.09      \\
    Spatial coverage area (m$^{2}$) &     30.38       &        53.46    \\
    Mean speed (m/s)                &    4.0975        &     3.5074  \\
    Speed standard deviation (m/s)  &    3.3707        &     2.8706  \\
    Max.\ speed (m/s)               &  19.1292          &     14.2290  \\
    Mean acceleration (m/s$^{2}$)   &   1.2640         &      1.0304      \\
    Acceleration standard deviation (m/s$^{2}$)   & 0.8814           & 0.8308           \\
    Max.\ acceleration (m/s$^{2}$)   &   6.6490         &   6.3652         \\
    Mean angle &-0.0286 & -0.6171\\
    Angle standard deviation &1.6033 & 2.0338\\
    Max.\ angle &3.1420 & 3.1420\\
    \hline
  \end{tabular}
\end{table}

\subsection{Metrics and Settings}
To evaluate the accuracy of predicted trajectories, we calculate Average Displacement Error (ADE) and Final Displacement Error (FDE) for baselines and our model in the next 3-second horizons. ADE and FDE can be computed by:
\begin{align}
\mathrm{ADE}&=\frac{1}{T_f} \sum_{t=1}^{T_f} \sqrt{\left(\hat{x}_{t}-x_{t}\right)^2+\left(\hat{y}_{t}-y_{t}\right)^2}\\
\mathrm{FDE}&=\sqrt{\left(\hat{x}_{T_f}-x_{T_f}\right)^2+\left(\hat{y}_{T_f}-y_{T_f}\right)^2}
\end{align}

Additionally, we compute the collision rate (CL) in the testing scenarios to evaluate the safety of the learned policy. 
Since our planner rolls out continuous trajectories and does not explicitly model a vehicle footprint, we determine a collision event purely based on pairwise trajectory proximity with a safety-distance threshold. 
Specifically, for each testing scene, we define the minimum inter-vehicle distance within the horizon as:
\begin{align}
    d_{\min} \;=\; \min_{t=1,\ldots,T_f}\;\min_{i<j}\; \left\lVert \mathbf p_i(t)-\mathbf p_j(t)\right\rVert_2
\end{align}

A collision is declared if $d_{\min}<d_{\mathrm{safe}}$, where $d_{\mathrm{safe}}$ is the same safety-distance threshold used in the safety term of Eq.~9 (set to $3$~m in our experiments). 
The collision rate is then computed at the scene level as:
\begin{align}
    \mathrm{CL} \;=\; \frac{1}{|\mathcal D_{\mathrm{test}}|}\sum_{m\in\mathcal D_{\mathrm{test}}}\mathbbm{1}\!\left[d_{\min}^{(m)}<d_{\mathrm{safe}}\right]
\end{align}
where $\mathcal D_{\mathrm{test}}$ is the set of testing scenes, $d_{\min}^{(m)}$ is the minimum inter-vehicle distance in scene $m$, and $\mathbbm{1}[\cdot]$ is the indicator function.

In this study, experiments are conducted on a server running Ubuntu 24.04.2, equipped with two NVIDIA RTX A6000 GPUs. The hidden dimension size is set to 64, and the learning rate is fixed at 1e-4. We employ the Theseus library \cite{pineda2022theseus} for differentiable nonlinear optimization of the potential function using the Levenberg–Marquardt algorithm, with a step size of 3e-1. After generating the planned motions, we use a unicycle model \cite{polack2017kinematic} to roll out the corresponding trajectories. The safety distance is set as 3 meters. Three parallel experiments are conducted.

\subsection{Baselines}
In this study, we compare our framework against several classical and state-of-the-art methods. 

\begin{itemize}
    \item IDM: Intelligent Driver Model (IDM) computes longitudinal acceleration based on the ego vehicle's current velocity and its distance to a leading vehicle. We follow the setting in \cite{chen2025dynamic} in our experiments.

    \item BC: Behavior Cloning \cite{farag2018behavior} employs supervised learning to imitate human driving behaviors. Given the historical trajectories of surrounding vehicles, the model predicts the future motion of the ego vehicle by maximizing the likelihood of observed actions.

    \item GAIL: Generative Adversarial Imitation Learning (GAIL) \cite{ho2016generative} follows the idea of Generative Adversarial Network by learning human-like driving policies. Here, we use PPO as the backbone network for training.

    \item GameFormer: GameFormer \cite{huang2023gameformer} is a Transformer-based model that is further trained using Level-K game-theoretic reasoning to capture interactive decision-making among agents.

    \item DIPP: Differentiable Integrated Prediction and Planning (DIPP) \cite{huang2023differentiable} adopts a Transformer-based architecture and incorporates a differentiable optimizer to perform motion planning in an end-to-end manner.

    \item EPO: As introduced in the \textit{Related Work}, EPO \cite{diehl2023energy} is an Energy-Based Potential Games (EPO) framework for vehicle trajectory prediction, but EPO trains the model by optimizing energy functions rather than iterative policy optimization.

    \item Diffuser: Diffuser is a data-driven planner based on modern diffusion models \cite{janner2022planning}, which formulates the planning problem as a conditional imputation task given the initial and goal states. It demonstrates strong performance in motion planning tasks. However, Diffuser does not explicitly model interactions among multiple agents and lacks theoretical guarantees.
    
\end{itemize}

\subsection{Overall Results}
Table~\ref{result} presents the overall results for the two scenarios. DFP-PDG achieves the lowest or second-lowest ADE and FDE in most cases, along with a zero collision rate. Notably, although our framework does not incorporate explicit safety constraints, the low collision rates result from its ability to accurately model authentic human driving behaviors. However, as demonstrated in \cite{chen2025dynamic}, a post-processing step can be readily applied to the planned motions to enforce safety constraints when necessary. 

Additionally, IDM exhibits the poorest performance, as it merely reacts to the behavior of the leading vehicle and fails to account for other surrounding vehicles. The relatively weak performance of traditional imitation learning methods further highlights that explicitly modeling interactions can substantially enhance human-likeness in driving behavior. Although Diffuser has been reported as one of the most advanced methods in imitation learning \cite{ubukata2024diffusion}, it conditions only on historical interactions and fails to explicitly model future interactions among agents. Moreover, GameFormer explicitly models human interactions using cognitive hierarchy theory and achieves performance comparable to our framework. EPO also achieves competitive ADE/FDE in both scenarios, suggesting that encoding interactions via an energy-based potential can yield strong point prediction accuracy.

We also present a comparison of the methods in Table~\ref{tab:method_comparison}, evaluating them in terms of interpretability, theoretical guarantees, and explicit interaction modeling. Our proposed method DFP-PDG is the only approach that simultaneously satisfies all three desired properties. Traditional imitation learning baselines such as BC and GAIL lack interpretability and offer no theoretical foundations, which limits their robustness and explainability. GameFormer, EPO and DIPP incorporate interaction-aware mechanisms, but they either lack theoretical grounding or rely on black-box diffusion models. Diffuser achieves strong empirical performance but does not provide interpretability or theoretical analysis. In contrast, DFP-PDG integrates a potential game structure and DFP, which not only models multi-agent interactions explicitly but also ensures interpretability and convergence guarantees through a theoretically grounded framework.

\begin{table*}[t]
\centering
\caption{Comparison Metrics in MA and GL Scenarios}
\review{
\label{result}
\renewcommand\arraystretch{1.2}
\resizebox{.9\linewidth}{!}{
\begin{tabular}{l|ccc|ccc}
\hline
\multirow{2}{*}{Model} & \multicolumn{3}{c|}{MA} & \multicolumn{3}{c}{GL} \\ \cline{2-7}
                 & ADE & FDE & CL & ADE & FDE & CL \\ \hline
IDM              &  4.6345&   8.2452   & 0.00\% &   3.9637  &  6.7603  &   0.00\%   \\ 
BC               & 0.4685$\pm$0.0513& 1.0504$\pm$0.1348   & 5.39\%$\pm$1.13\%&    0.5783$\pm$0.0919& 1.1583$\pm$0.1534  & 5.13\%$\pm$0.87\%     \\ 
GAIL             & 0.4206$\pm$0.0393 & 0.9477$\pm$0.1379 &  3.26\%$\pm$0.68\%&   0.3798$\pm$0.0257 & 0.8589$\pm$0.1584  & 5.05\%$\pm$1.66\%    \\ 
GameFormer       &  0.2631$\pm$0.0480&   0.3876$\pm$0.0121& 0.00\%$\pm$0.00\%&  0.2653$\pm$0.0156  & 0.3481$\pm$0.0065  & 0.00\%$\pm$0.00\%    \\ 
DIPP             &   0.3067$\pm$0.0011&0.5925$\pm$0.0045&0.26\%$\pm$0.05\%& 0.3733$\pm$0.0063&  0.5397$\pm$0.0063 & 0.22\%$\pm$0.11\%    \\ 
EPO & 0.2679$\pm$0.0052 & 0.4313$\pm$0.0052 &  0.00\%$\pm$0.00\% &0.2555$\pm$0.0088& 0.3969$\pm$0.0109&0.00\%$\pm$0.00\%\\
Diffuser     & 0.2718$\pm$0.0021&  0.3754$\pm$0.0064   &0.24\%$\pm$0.12\% &  0.2860$\pm$0.0083     & 0.4181$\pm$0.0032    & 0.00\%$\pm$0.00\%    \\ \hline
DFP-PDG & 0.2475$\pm$0.0068& 0.3668$\pm$0.0090& 0.00\%$\pm$0.00\%&0.2603$\pm$0.0062&0.3651$\pm$0.0081&0.00\%$\pm$0.00\%\\ \hline
\end{tabular}}}
\end{table*}

\begin{table*}[h]
\centering
\caption{Comparison of Methods: Interpretability, Theoretical Guarantee, and Interaction Modeling}
\label{tab:method_comparison}
\renewcommand\arraystretch{1.3}
\begin{tabular}{lccc}
\hline
\textbf{Method} & \textbf{Interpretability} & \textbf{Theoretical Guarantee} & \textbf{Explicit Interaction Modeling} \\
\hline
BC          & \xmark  & \xmark         & \xmark \\
GAIL       & \xmark  & \xmark         & \xmark \\
GameFormer   & \xmark  & \xmark         & \checkmark \\
DIPP      & \checkmark  & \xmark     & \checkmark \\
EPO & \checkmark  & \xmark     & \checkmark\\
Diffuser & \xmark & \xmark &\xmark \\
DFP-PDG & \checkmark  & \checkmark     & \checkmark \\
\hline
\end{tabular}
\end{table*}

\subsection{Weight Interpretation}
\review{One advantage of our framework is that the learned weights offer interpretable insights into human driving behaviors. Table \ref{tab:weight} presents the learned global weights $\lambda_{(\cdot)}$. With all initial weights set to 1, the learned weights remain close to this value. This is because all cost terms are already normalized to a comparable small scale, effectively achieving a near-optimal trade-off among the four components and stabilizing the nonlinear solver. Moreover, the model is more sensitive to the relative proportions of the sub-losses than to their absolute magnitudes. Once these proportions are balanced, deviating any single weight from 1 yields no additional benefit, leading the parameters to naturally converge near their initialization.}

\begin{table}[h]
\centering
\caption{Learned Global Weights and Ranges for MA and GL Scenarios}
\label{tab:weight}
\renewcommand\arraystretch{1.3}
\review{
\begin{tabular}{lcc}
\hline
\textbf{Weight Term} & \textbf{MA} & \textbf{GL} \\
\hline
$\lambda_{\text{goal}}$       &     0.9915  & 0.9914 \\
$\lambda_{\text{smooth}}$     &     1.0059  & 1.0190  \\
$\lambda_{\text{efficiency}}$ &      1.0068  &0.9925   \\
$\lambda_{\text{safety}}$     &     0.9984   & 0.9989 \\
\hline
\end{tabular}}
\end{table}

\begin{figure*}[htbp]
  \centering
  \includegraphics[width=.85\linewidth]{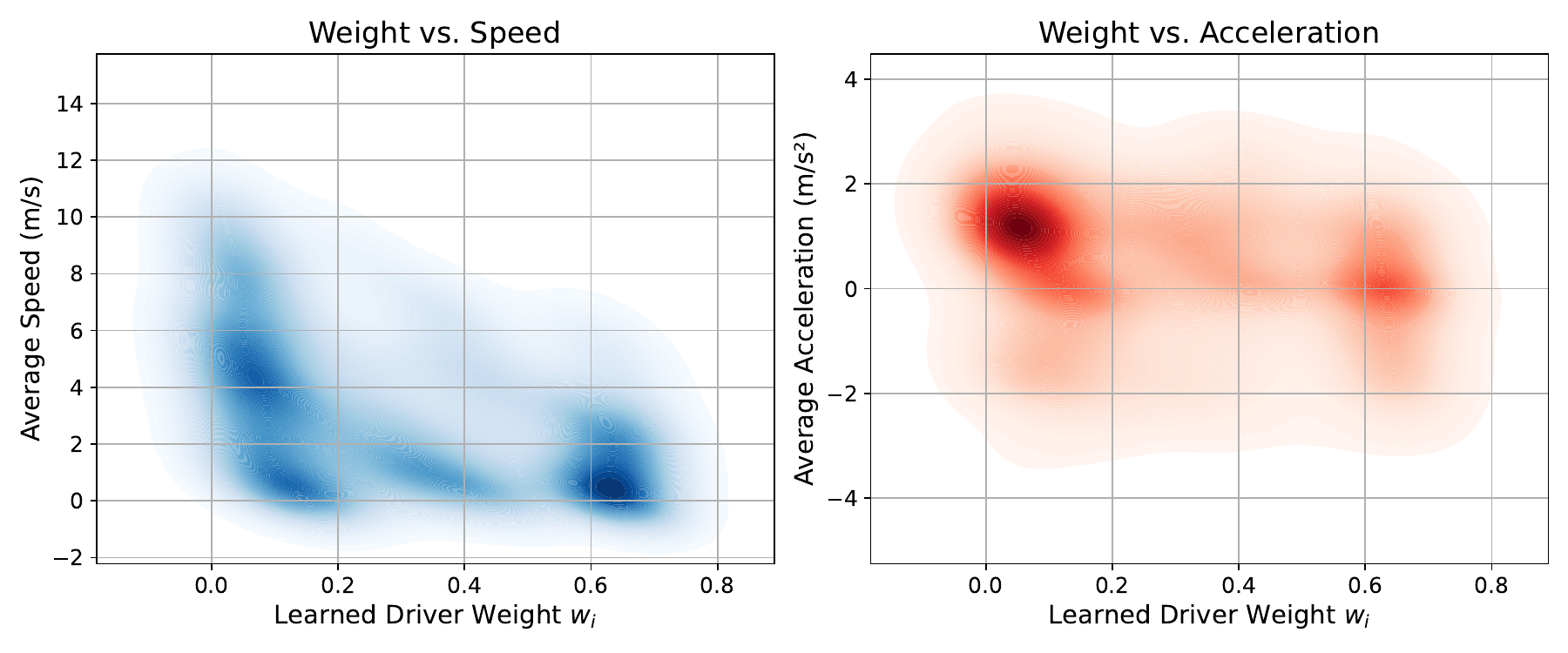}
  \caption{\review{Density plots of learned weights versus speed and acceleration, computed based on 1-second historical trajectories. These plots illustrate how driving style sensitivity (as represented by the weight) correlates with vehicle dynamics.}}
  \label{fig:dist}
\end{figure*}

\review{Fig.~\ref{fig:dist} presents the learned individual weights \(w_i\) and the corresponding driving statistics using the MA scenario as an example. The results show that higher-speed trajectories are mainly associated with smaller weights, whereas larger weights are concentrated around low-speed trajectories. The acceleration plot shows a consistent pattern, where larger positive accelerations are more frequently observed in the low-weight region, while high-weight samples are mostly associated with mild accelerations. These observations support the interpretation of \(w_i\) as an individual safety-sensitivity factor: smaller weights indicate lower sensitivity to safety-related cost changes and more aggressive driving behavior, whereas larger weights correspond to more conservative driving profiles. }

\subsection{Ablation Study}
To demonstrate the effectiveness of each component, we conduct an ablation study in this section. Specifically, we compare the original DFP-PDG framework with the following variants:
(i) DFP-PDG$^{\text{-IW}}$ – removes individual weights, only using learnable global weighting instead;
(ii) DFP-PDG$^{\text{-IR}}$ – removes the interaction-related modules in the deep policy network and merely uses information of the ego vehicle;
(iii) DFP-PDG$^{\text{-SC}}$ – removes the efficiency and smoothness terms, retaining only the goal attraction and safety components.

As shown in Table \ref{tab:ablation}, each component of the DFP-PDG framework contributes to the final performance. Removing the individualized weighting mechanism (DFP-PDG\textsuperscript{-IW}) leads to a mild performance drop in both MA and GL scenarios, indicating the importance of personalized behavior modeling. Eliminating the interaction-related modules (DFP-PDG\textsuperscript{-IR}) also results in considerable degradation, confirming that accounting for neighboring vehicles is crucial at unsignalized intersections. Furthermore, removing the efficiency and smoothness terms (DFP-PDG\textsuperscript{-SC}) slightly degrades the results, suggesting that these terms contribute to fine-tuning motion quality, but are less critical compared to the goal and safety objectives. 

\begin{table}[h]
\centering
\caption{Ablation Study of the DFP-PDG Framework}
\label{tab:ablation}
\review{
\renewcommand\arraystretch{1.3}
% \resizebox{\textwidth}{!}{
\begin{tabular}{l|cc|cc}
\hline
\multirow{2}{*}{\textbf{Model Variant}} & \multicolumn{2}{c|}{\textbf{MA}} & \multicolumn{2}{c}{\textbf{GL}} \\
 & ADE  & FDE  & ADE  & FDE \\ \hline
DFP-PDG             &   0.2475     &    0.3668    &   0.2603     &   0.3651              \\
\quad - IW     &   0.2592     &   0.4107     &   0.2807     &  0.4703           \\
\quad - IR    &   0.4245     &  0.7682      &   0.3555     &   0.6586               \\
\quad - SC  &   0.2707     &   0.4414     &  0.2856    & 0.4660        \\ \hline
\end{tabular}}
\end{table}

\subsection{Computational Efficiency}
Inference time is critical for planning and prediction, both in simulation and in real-world autonomous driving deployment. In this section, we compare the test-time computational efficiency of different models. While training DFP–PDG is relatively slow due to the alternating best-response iterations, inference is efficient: at test time, we can directly deploy the trained deep policy for motion planning.

As shown in Table~\ref{tab:runtime_params}, we compare inference efficiency across all baselines using per-agent runtime and learnable parameters. As a classical baseline, IDM is the simplest model and therefore has the fastest inference time. 
Simple policy-only baselines (i.e. BC and GAIL) are the most efficient among learning-based methods, requiring only a single forward pass and thus achieving $\sim$18~ms/agent with $\sim$0.45--0.55M parameters. In contrast, methods that rely on iterative refinement or sampling are substantially slower. DIPP is the most expensive baseline in our comparison ($\sim$300~ms/agent) because it embeds a differentiable nonlinear optimizer and typically requires many iterations at test time. Similarly, diffusion-based planning incurs high latency ($\sim$160~ms/agent) due to repeated denoising steps. EPO also introduces additional computational overhead ($\sim$102~ms/agent) from solving multiple energy-minimization modes and unrolling the optimizer. 

Our DFP-PDG achieves a favorable trade-off between efficiency and modeling capacity: it runs in $\sim$68~ms/agent while using only $\sim$0.45M learnable parameters, more efficient than optimization-heavy baselines (DIPP/EPO) and sampling-heavy diffusion planners. GameFormer exhibits a larger model size ($\sim$3.62M parameters) due to the stack of Transformer layers and comparable runtime ($\sim$67~ms/agent), suggesting that our approach attains similar inference speed with a significantly smaller network by exploiting the potential-game structure and structured best-response updates. Overall, these results highlight that explicitly leveraging game structure can reduce model size and avoid excessive test-time optimization/sampling, leading to a more practical inference profile for interactive driving.

\begin{table}[t]
\centering
\caption{Runtime and learnable parameters comparison.}
\label{tab:runtime_params}
\begin{tabular}{lcc}
\toprule
Model & Runtime (ms/agent) & Learnable Params \\
\midrule
IDM & 2 & $\sim$0\\
BC & 17 & 0.45 M\\
GAIL & 18 & 0.55 M\\
GameFormer & 67 & 3.62 M \\
DIPP & 300 & 0.46 M\\
EPO & 102 &  0.62 M \\
Diffuser & 160 & 0.50 M\\
DFP-PDG & 68 & 0.45 M \\
\bottomrule
\end{tabular}
\end{table}

\subsection{Quantitative Visualization}
\begin{figure*}[htbp]
  \centering
  \includegraphics[width=\linewidth]{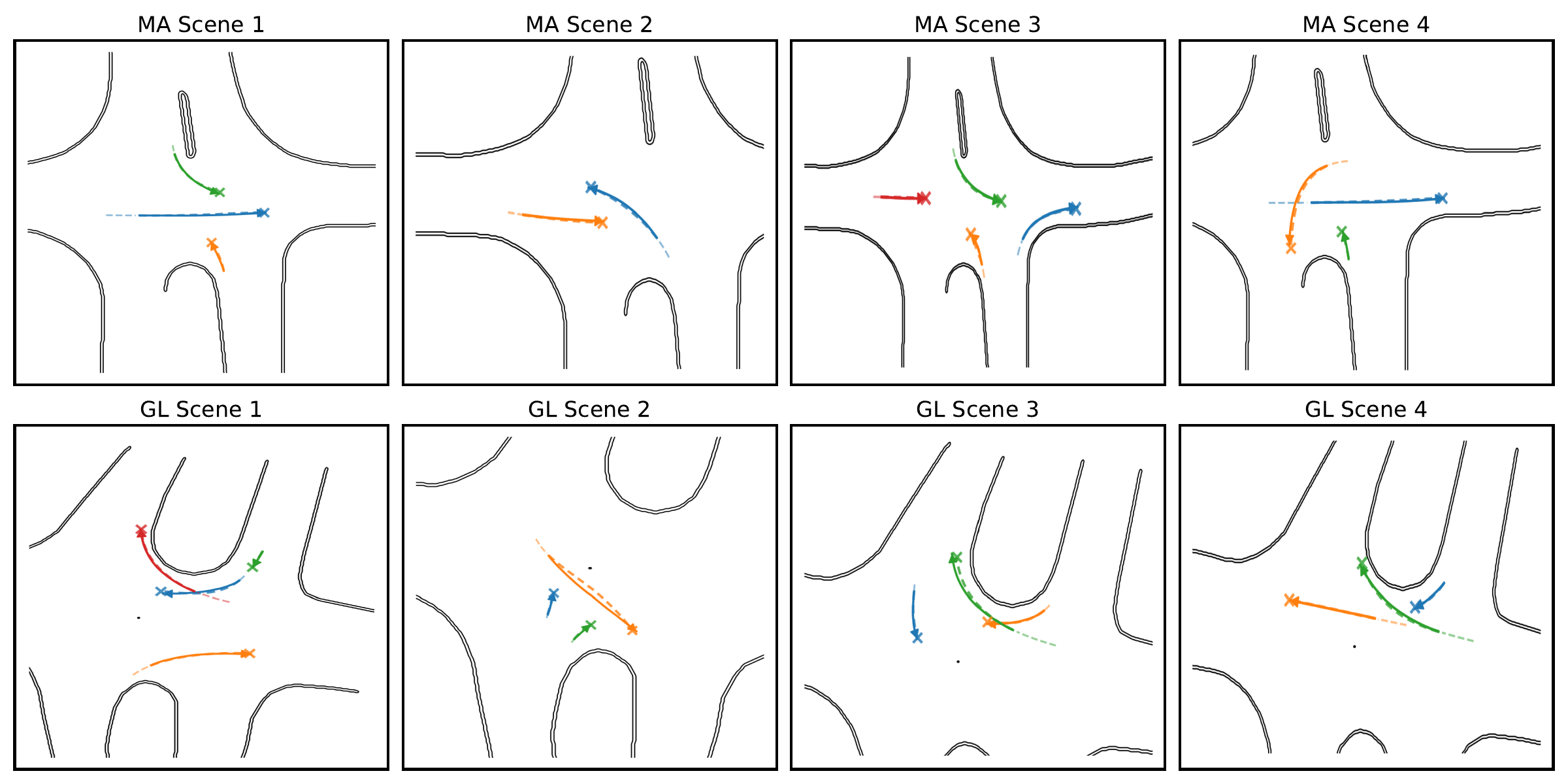}
  \caption{Case visualization of the planned trajectories. Light and dark dashed lines represent the observed historical and ground truth trajectories, respectively. Solid lines indicate the planned trajectories. Cross marks denote the true goal positions, while arrows represent the planned destinations.}
  \label{fig:case}
\end{figure*}

In this section, we randomly select several cases involving different agents to visualize the planning process. In Fig.~\ref{fig:case}, the light and dark dashed lines represent the observed historical and ground truth trajectories, respectively. Solid lines indicate the planned trajectories, with cross marks denoting the true goal positions and arrows representing the planned destinations. As illustrated in the figure, the proposed DFP-PDG framework is capable of accurately planning motions in highly interactive scenarios.

\subsection{Trajectory Prediction with DFP-PDG}   
The discussions and results presented thus far have been conditioned on the goal state, framing the task as motion planning. In this section, we demonstrate that the DFP-PDG framework can also be directly applied to trajectory prediction by eliminating goal states and goal cost. We also conduct comparative experiments; note that DIPP is excluded from the comparison as it inherently requires goal state information.

As shown in Table \ref{pred_res}, the proposed DFP-PDG framework still achieves good performance in both MA and GL scenarios across all metrics. The absence of goal state information leads to a significant decline in performance. Notably, the collision rate is still zero for DFP-PDG, indicating that our method inherently avoids unsafe interactions with the implementation of the non-linear optimizer. Compared to state-of-the-art predictors such as Diffuser and GameFormer, DFP-PDG not only yields comparable trajectory errors but also ensures safer predictions.

\begin{table*}[t]
\centering
\caption{Predictor Performance in MA and GL Scenarios}
\label{pred_res}
\review{
\renewcommand\arraystretch{1.2}
\resizebox{0.6\linewidth}{!}{
\begin{tabular}{l|ccc|ccc}
\hline
\multirow{2}{*}{Model} & \multicolumn{3}{c|}{MA} & \multicolumn{3}{c}{GL} \\ \cline{2-7}
                 & ADE & FDE & CL & ADE & FDE & CL \\ \hline
IDM              &  4.6345&   8.2452   & 0.00\% &   3.9637  &  6.7603  &   0.00\%   \\ 
BC               & 1.5507 & 2.6979&11.42\%& 1.5406&2.6920 &9.78\% \\ 
GAIL             &  1.4420&2.5291&5.54\% & 1.6723  & 2.8623 &6.74\% \\ 
GameFormer       &  0.7839 & 1.6358 & 0.53\%  & 0.8308  & 1.6623 & 1.02\%  \\ 
EPO & 0.8126& 1.6197&0.00\% & 0.7966& 1.6209&0.00\% \\
Diffuser     & 1.1349  & 2.1073& 2.77\%& 1.1016&2.0544 & 3.13\% \\ \hline
DFP-PDG & 0.8433 & 1.6813 & 0.00\% &   0.7742  &  1.5334   &0.00\% \\ \hline
\end{tabular}
}}
\end{table*}

\section{Conclusion}
In this study, we propose a novel framework, Deep Fictitious Play-Based Potential Differential Game (DFP-PDG), to model complex interactions among vehicles. The differential game is reformulated as a weighted potential game, and the driving policy is initially learned through a deep neural network and subsequently refined using differentiable nonlinear optimization. To the best of our knowledge, this is the first study to apply Deep Fictitious Play (DFP) to multi-vehicle interaction modeling with theoretical guarantees. Extensive experiments on the INTERACTION MA and GL scenarios demonstrate that DFP-PDG achieves state-of-the-art ADE/FDE accuracy while maintaining a 0\% collision rate, validating both its safety and effectiveness. The learned weights further provide an interpretable link between driving styles and trajectory aggressiveness, offering insight beyond black-box data-driven planners.

Compared with existing methods, the proposed framework supplies a rigorous game-theoretic foundation with a proven fictitious-play convergence guarantee under stated assumptions and couples semantic graph encoding, potential-based objectives, and differentiable Levenberg–Marquardt optimization in an end-to-end trainable manner. These qualities collectively advance the state of interactive motion planning toward explainable, theoretically grounded, and data-efficient autonomous driving.

However, DFP-PDG still incurs high computational complexity in the training stage due to the use of cyclic best-response dynamics. In addition, while we incorporate smoothness- and safety-related terms in the training objective to encourage physically plausible and collision-averse interactions, it is important to note that these terms are currently implemented as soft constraints: they penalize undesirable behaviors in the loss, but do not provide deterministic feasibility or safety guarantees in all scenarios. For instance, a few of the predicted actions ($<$1\%) still exhibit relatively large magnitudes. 
This limitation is not unique to our approach; it is a common issue across learning-based trajectory generation/prediction methods, where models may capture favorable average-case statistics yet still exhibit local violations of smoothness or safety boundaries in long-tail, noisy, or out-of-distribution cases \cite{hagenus2024survey}. Importantly, our framework is naturally extensible toward stronger guarantees. On the one hand, critical constraints (e.g., collision avoidance, and bounds on control and control-rate) can be enforced as hard constraints by embedding them into a differentiable optimization/planning layer \cite{amos2018differentiable, yang2022differentiable}. On the other hand, a lightweight post-processing or safety-filtering stage (e.g., constraint-based trajectory repair, control projection, or a safety monitor) can be applied at inference time to correct constraint violations \cite{chen2025dynamic}. These extensions enable us to preserve the learned interactive behaviors and diversity while explicitly reducing smoothness and safety violations.

% use section* for acknowledgment
\section*{Acknowledgment}
We thank the Pacific Northwest's Transportation Consortium (PacTrans) Center for their funding support.

% Can use something like this to put references on a page
% by themselves when using endfloat and the captionsoff option.
\ifCLASSOPTIONcaptionsoff
  \newpage
\fi

\bibliographystyle{IEEEtran}
\bibliography{refs}

% insert where needed to balance the two columns on the last page with
% biographies
%\newpage
\vspace{11pt}

\begin{IEEEbiography}[{\includegraphics[width=1in,height=1.35in,clip,keepaspectratio]{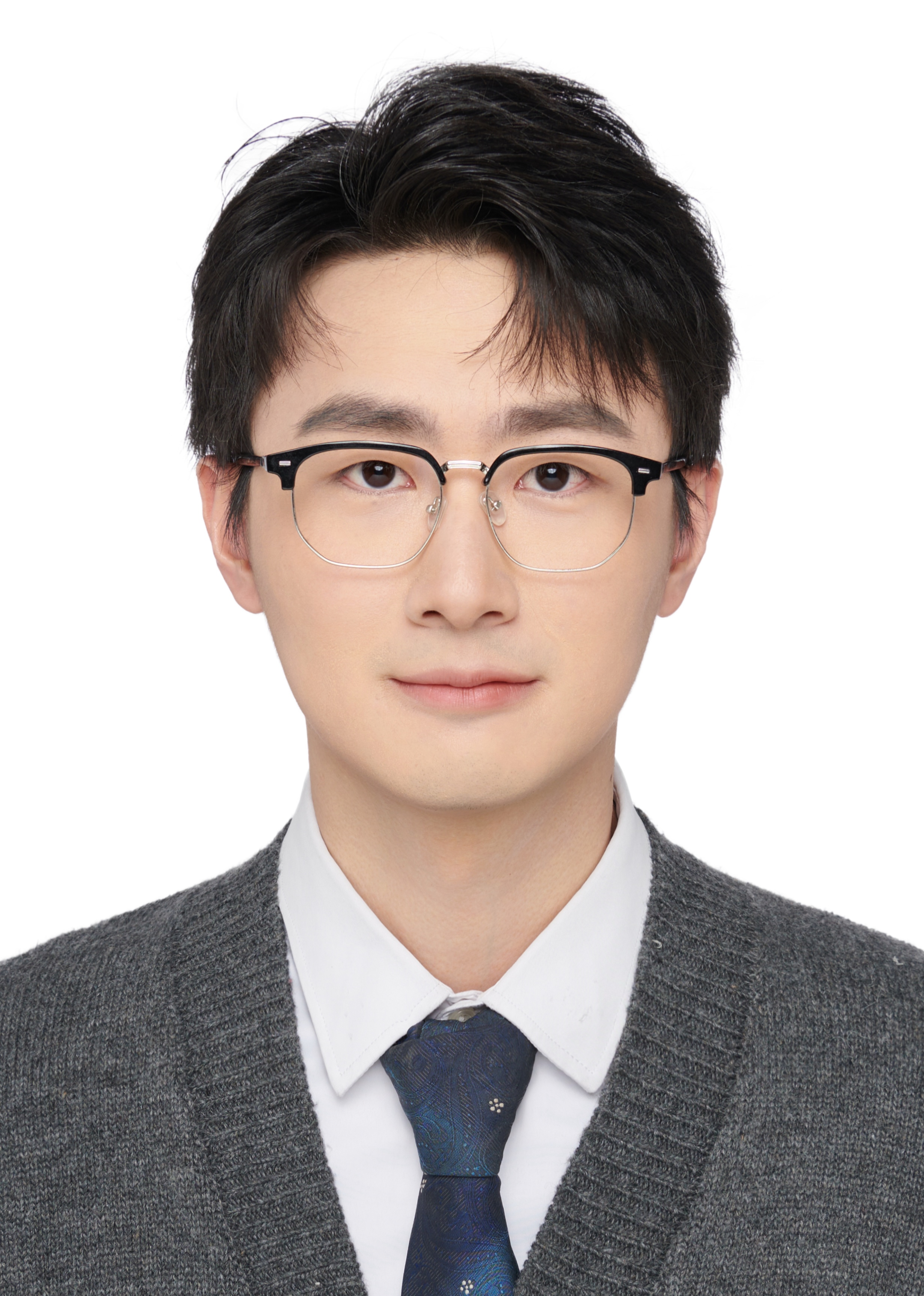}}]{Kehua Chen} received a B.S. degree in Civil Engineering from Chongqing University and a dual M.S. degree in Environmental Sciences from the University of Chinese Academy of Sciences and the University of Copenhagen. He earned his Ph.D. in Intelligent Transportation from the Hong Kong University of Science and Technology in 2024. Currently, he is a postdoctoral scholar at the Smart Transportation Applications and Research (STAR) Lab at the University of Washington. His research interests encompass urban and sustainable computing, as well as autonomous driving.
\end{IEEEbiography} 

\begin{IEEEbiography}
[{\includegraphics[width=1in,height=1.35in,clip,keepaspectratio]{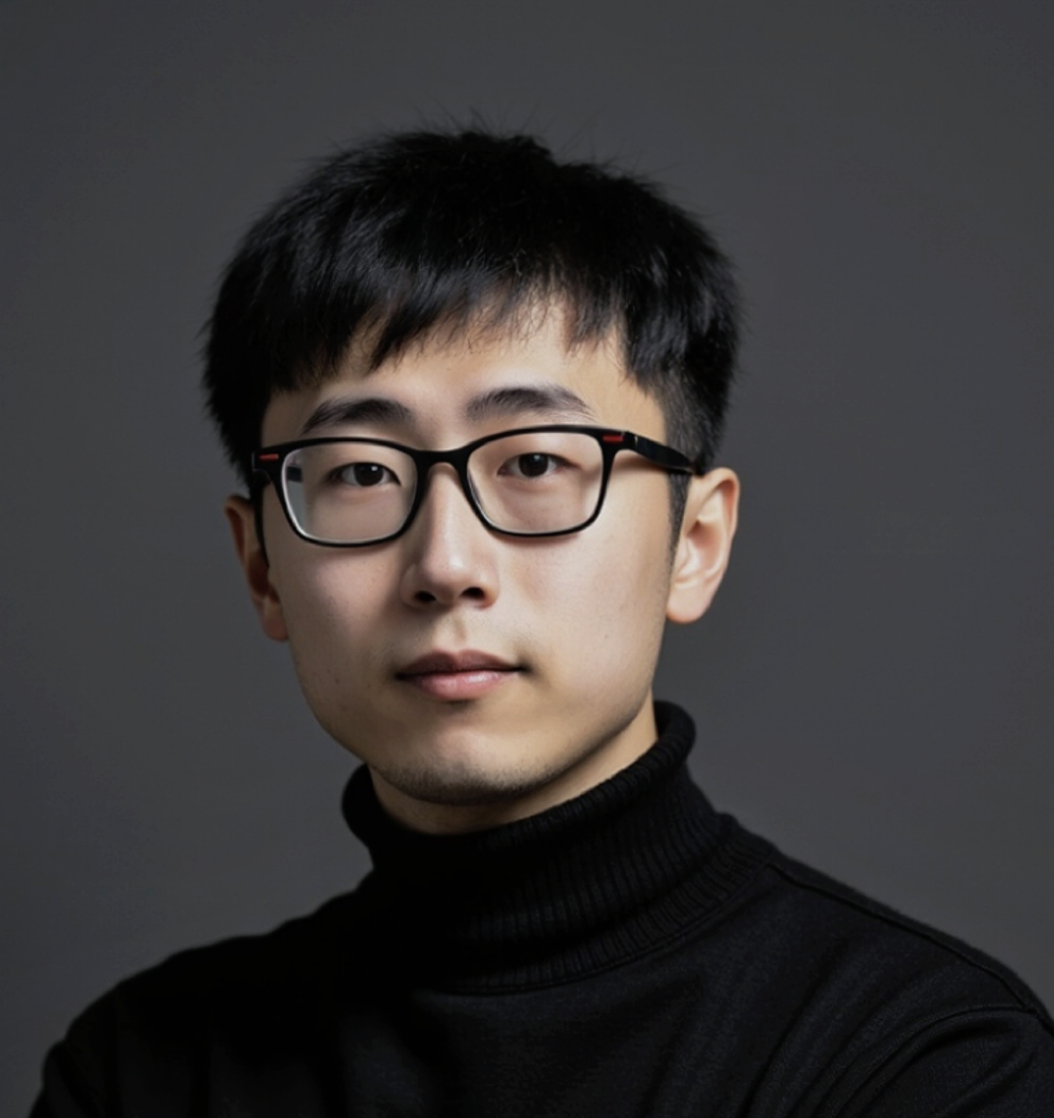}}]
{Ryan Feng Lin} is a Ph.D. candidate in the Department of Industrial \& Systems Engineering at the University of Washington. He received his M.S. in Computer Science and dual B.S. degrees in Statistics and Computer Science from the University of Science and Technology of China. His research lies at the intersection of AI/machine learning, statistics, and quality science, focusing on application-driven methodological development for building personalized digital twins, with applications in healthcare, transportation, and marketing. 
\end{IEEEbiography}

\begin{IEEEbiography}[{\includegraphics[width=1in,height=1.35in,clip,keepaspectratio]{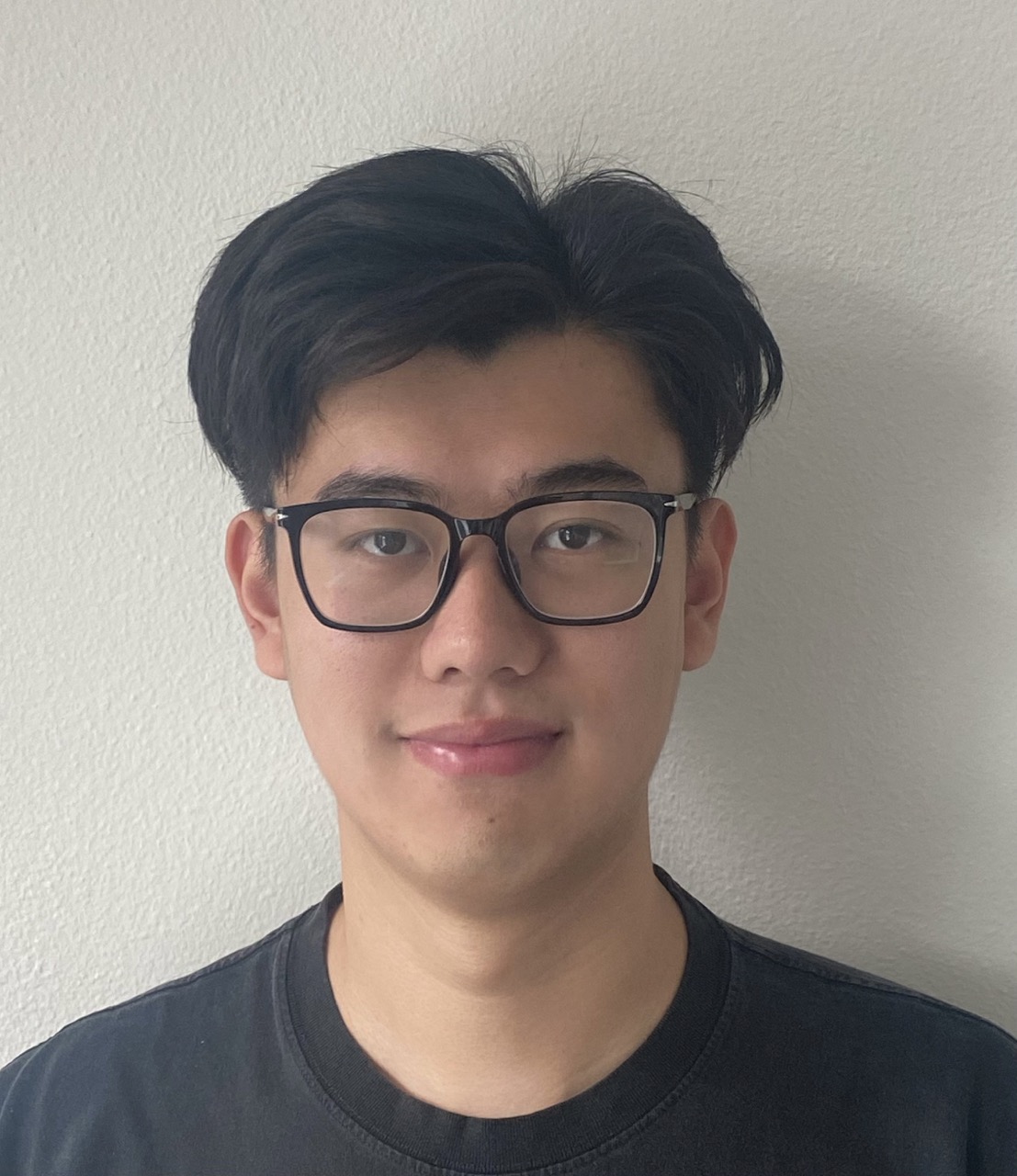}}]{Shucheng Zhang}
\normalfont is currently pursuing the Ph.D. degree in the Smart Transportation Application and Research (STAR) Lab, Department of Civil and Environmental Engineering at the University of Washington. Prior to joining the STAR Lab, Shucheng earned his M.S. in Mechanical Engineering from Duke University. His research interests include computer vision, intelligent transportation systems, and autonomous vehicles, with a focus on developing innovative solutions to enhance road safety and transportation automation.
\end{IEEEbiography}

\begin{IEEEbiography}[{\includegraphics[width=1in,height=1.35in,clip,keepaspectratio]{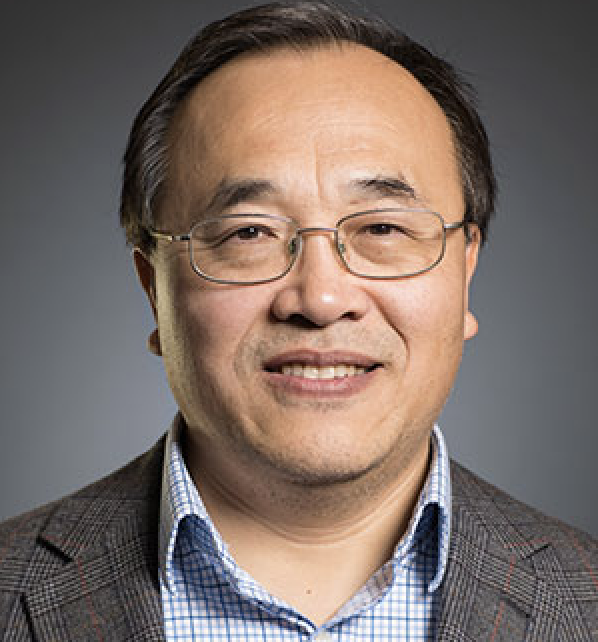}}]{Yinhai Wang}
\normalfont received the master’s degree in computer science from the University of Washington (UW) and the Ph.D. degree in transportation engineering from The University of Tokyo in 1998. He is currently a Professor in transportation engineering and the Founding Director of the Smart Transportation Applications and Research Laboratory (STAR Lab), UW. He also serves as the Director of the Pacific Northwest Transportation Consortium (PacTrans), U.S. Department of Transportation, University Transportation Center for Federal Region 10.
\end{IEEEbiography}

% that's all folks

\end{document}